\def\ARXIV{1} 

\documentclass{article}

\PassOptionsToPackage{numbers, compress}{natbib}
\usepackage[eandd, final]{neurips_2026}

\usepackage[utf8]{inputenc}
\usepackage[T1]{fontenc}
\usepackage{hyperref}
\usepackage{url}
\usepackage{booktabs}
\usepackage{amsfonts}
\usepackage{amsmath}
\usepackage{nicefrac}
\usepackage{microtype}
\usepackage{graphicx}
\usepackage{xcolor}
\usepackage{colortbl}
\usepackage{multirow}
\usepackage{makecell}
\usepackage{subcaption}
\usepackage{array}
\usepackage{tabularx}
\hypersetup{hidelinks}

\newcommand{\best}[1]{\textcolor{red}{\textbf{#1}}}
\newcommand{\second}[1]{\textcolor{blue}{\textbf{#1}}}
\definecolor{tablehead}{gray}{0.92}
\definecolor{tablesubhead}{gray}{0.96}

\title{StakeBench: Evaluating Language Understanding Grounded in Market Commitment}

\ifnum\ARXIV=1 
    \usepackage{orcidlink}
    \hypersetup{hidelinks}
    
    \author{%
      Yunhua Pei\thanks{Corresponding author: \texttt{ge22472@bristol.ac.uk}} \;\orcidlink{0000-0003-2906-0827} \ 
      Jingyu Hu\,\orcidlink{0000-0002-4566-5327} \ 
      Yiwei Shi\,\orcidlink{0000-0001-6522-9875} \ 
      Hongnan Ma\,\orcidlink{0009-0004-5767-4089} \  
      Weiru Liu\,\orcidlink{0000-0001-8356-1361} \  
      John Cartlidge\,\orcidlink{0000-0002-3143-6355}\\
      University of Bristol, Bristol, UK \\
      \texttt{\{ge22472,jingyu.hu,yiwei.shi\}@bristol.ac.uk} \\
      \texttt{\{hongnan.ma,weiru.liu,john.cartlidge\}@bristol.ac.uk}
    }
    \usepackage{fancyhdr}
    \usepackage{lastpage}
    \fancyheadoffset[R]{-0.0cm}
    
\else
    \author{%
      Yunhua Pei \quad Jingyu Hu \quad Yiwei Shi \quad Hongnan Ma \quad Weiru Liu \quad John Cartlidge \\
      University of Bristol, Bristol, UK \\
      \texttt{\{ge22472,jingyu.hu,yiwei.shi\}@bristol.ac.uk} \\
      \texttt{\{hongnan.ma,weiru.liu,john.cartlidge\}@bristol.ac.uk}
    }
\fi

\begin{document}

\maketitle

\ifnum\ARXIV=1
    \setcounter{footnote}{0} 
\fi

\begin{abstract}
Existing financial NLP benchmarks often rely on labels supplied by outside observers, measuring how language is perceived rather than what speakers have committed to in the market. We introduce \textbf{StakeBench}, an evaluation framework for language understanding grounded in market commitment. StakeBench links 560,876 comments from 2,261 resolved markets to verified position, action, and market-odds records across Polymarket and Manifold. Supervision is derived from observable market behavior. Position sides, post-comment trading actions, and market-odds trajectories replace human annotation. Four diagnostic tasks test whether models detect market commitment, identify the revealed side, anticipate future action, and perform collective odds projection. Three commitment-aware metrics measure alignment with revealed preferences rather than perceived sentiment. Validity audits and explicit interpretation boundaries help distinguish observable commitment signals from latent belief and causal market-odds impact. Across 15 LLMs and 18 topics and platform settings, models partially recover position-side signals, with Directed Accuracy from 0.506 to 0.599, but show structural failures on later tasks. Ten of the fifteen models collapse to one or two action labels in future action anticipation, and no model consistently improves on the naive odds-direction baseline in collective odds projection. Model scale is not correlated with performance, finance-domain tuning does not improve revealed-side identification, and platform incentives strongly shape higher-order results. StakeBench is packaged with evaluation code\footnote{\url{https://github.com/PEIYUNHUA/stakebench_code_NeurIPS26}} and dataset\footnote{\url{https://huggingface.co/datasets/greenday123/StakeBench}} under CC-BY~4.0.
\end{abstract}

\section{Introduction}
\label{sec:intro}

Large language models (LLMs) have shown strong potential for financial text analysis, forecasting, and decision-making~\citep{yang2020,xie2024}. However, language in financial markets is not merely descriptive. Market participants often communicate while holding positions, facing reputational costs, or attempting to influence others' beliefs. This makes financial language intrinsically strategic: the same statement can carry different implications depending on what the speaker has at stake.
For example, online discussions were tightly associated with the GameStop short squeeze in 2021, while social-platform coordination has been linked to rapid pump-and-dump movements in cryptocurrency markets~\citep{buz2024,lamorgia2023}. These cases suggest a central challenge for financial NLP: understanding market language requires reasoning not only about what is said, but also about what the speaker has observably committed to.

Existing financial NLP resources cover text-level tasks from sentiment analysis
and stance detection to question answering and outcome
forecasting~\citep{yang2020,xie2024,xie2023pixiu,zhang2024reality,mohammad2016},
yet their supervision is defined by task labels, benchmark annotations, or
downstream outcomes rather than speaker commitment.
FinBen~\citep{xie2024} and PIXIU~\citep{xie2023pixiu} together cover broad
financial NLP task suites, and \citet{zhang2024reality} extend sentiment
evaluation across 26 datasets, but these targets still describe how text is
classified rather than what the speaker demonstrably chose in the market.
SemEval-2016 Task~6~\citep{mohammad2016} defines tweet-target stance
classification with three labels (FAVOR/AGAINST/NEITHER), where systems
infer whether a tweet supports or opposes a specified target. This label structure is closest to StakeBench, but the data consists of crowdsourced tweets rather than comments linked to verifiable financial positions.
Social-media analyses of investment communities~\citep{kheiri2023llms,buz2024}
analyze Reddit sentiment and investment advice without linking individual posts
to their authors' actual market positions or subsequent trading actions.
On the prediction-market side, Autocast~\citep{zou2022autocast},
\citet{halawi2024}, and ForecastBench~\citep{karger2025} benchmark LLM
forecasting calibration on resolved outcomes, targeting event resolution rather
than the commitment signals embedded in individual comments, while
\citet{turtel2025} use prediction-market-style forecasting data for model
self-improvement but do not evaluate comment text.
No benchmark jointly connects a speaker's language to their verified financial
position, subsequent trading behavior, and the collective market-odds signals
their speech may anticipate.

To bridge this gap, we introduce \textbf{StakeBench}, an evaluation framework
for commitment-grounded language understanding in prediction markets.
The key idea is to replace perception-based supervision with revealed-preference
signals derived from observable market commitments~\citep{samuelson1948}.
Prediction markets provide a natural testbed because comments, positions, trading
actions, resolutions, and market-odds trajectories can be jointly observed.
StakeBench contains 560,876 comments from 2,261 resolved markets across Polymarket and Manifold platforms.
This dual-platform design captures complementary incentive regimes: Polymarket
offers stronger monetary commitment signals, while Manifold provides broader
position coverage and direct resolution metadata.
Because these targets are behavioral proxies, StakeBench evaluates whether
models recover observable commitments rather than private beliefs or causal
effects of language on market odds. Full related works, see Appendix~\ref{sec:related}.

\begin{figure}[t]
    \centering
    \includegraphics[width=1\linewidth]{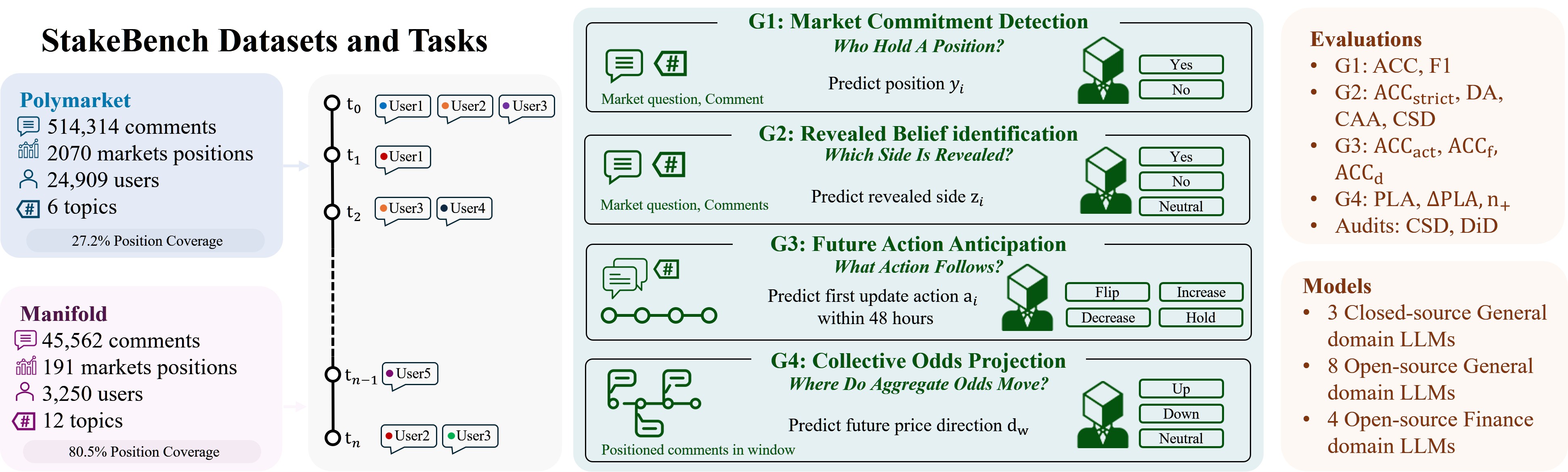}
    \caption{StakeBench overview. Comments from Polymarket and Manifold are linked to position, action, and market-odds records for four grounding tasks: market commitment detection, revealed-side identification, future action anticipation, and collective odds projection.}
    \label{fig:stakebench_overview}
\end{figure}

StakeBench is organized as a progressive diagnostic suite with four grounding
tasks, denoted \(\mathsf{G}_1\)--\(\mathsf{G}_4\).
The four tasks pair technical names with plain-language questions:
\textbf{\(\mathsf{G}_1\): Market Commitment Detection (Who Holds a Position?)}
asks whether a given comment is written by a user who holds any observable
position in the corresponding market.
\textbf{\(\mathsf{G}_2\): Revealed-Side Identification (Which Side Is Revealed?)}
asks which market side the commenter's revealed position supports.
\textbf{\(\mathsf{G}_3\): Future Action Anticipation (What Action Follows?)}
asks how the user will update the position after commenting.
\textbf{\(\mathsf{G}_4\): Collective Odds Projection (Where Do Aggregate Odds Move?)}
moves to the collective level, testing whether aggregated commitment-aware
comment signals project the next-window market odds direction.
The four tasks form an increasingly demanding hierarchy: detecting whether a
comment reflects market commitment, identifying the revealed side,
anticipating the user's subsequent action, and projecting whether collective
committed speech leads future odds movement.

Beyond prediction markets, StakeBench instantiates a general NLP problem: inferring a speaker's latent commitment from language when that commitment is later observable. Three uses follow. Position-aware auditing of financial communication asks whether a speaker is talking their own book, the primitive behind manipulation detection and disclosure compliance. Annotation-free evaluation applies wherever commitment records are public, subject to venue-specific validation, where most financial NLP resources still rely on
annotator-perceived sentiment~\citep{malo2014good}. Finally, since
finance-domain tuning does not improve speaker-level reasoning in our results, position-conditioned training objectives are a natural next target. In terms of \citet{crawford1982} and \citet{spence1978job}, the benchmark asks whether a model can tell costly commitment from cheap talk when the cost is observable and the outcome later settles.

In summary, our main contributions are:
1) We introduce \textbf{StakeBench}, a dual-platform prediction-market evaluation
framework with 560,876 comments from 2,261 resolved markets, linking language
to speaker positions, later actions, and market-odds histories.
2) We derive task targets from public position records, post-comment trading
behavior, market resolutions, and market-odds trajectories, replacing
perception-based labels with observable commitment signals.
3) We define four progressive diagnostic tasks covering market commitment
detection, revealed-side identification, future action anticipation, and
collective odds projection,
enabling evaluation beyond text-only sentiment or stance classification.
4) We provide commitment-aware metrics, validity audits, and documented
interpretation boundaries that support reusable evaluation of current and
future models.


\section{StakeBench}
\label{sec:dataset}

The aim of StakeBench is to link prediction-market comments to each author's verified
position record across resolved markets, creating four annotation-free evaluation tasks spanning the language-position-action-odds chain.
Specifically, StakeBench contains a total of 560,876 comments across 2,261 resolved-market records with 18
topic--platform combinations collected from two public prediction-market platforms:
\textbf{Polymarket},\footnote{\url{https://polymarket.com}} a real-money market whose
comments provide monetary commitment signals, and \textbf{Manifold},\footnote{\url{https://manifold.markets}} a virtual-Mana market with denser position coverage and direct resolved-outcome metadata.
For Polymarket, we retrieve comment threads via the Gamma application programming interface (API) and reconstruct each commenter's position at comment time by replaying their trade history against the activity API to compute \texttt{position\_side} (YES/NO) and \texttt{position\_size}.
For Manifold, positions are reconstructed from per-user bet records. \(\mathsf{G}_3\) additionally uses 862,540 Polymarket and 387,263 Manifold timestamped position-change records.

Markets and comments pass quality thresholds.
All data are retrieved from public APIs without authentication.
Table~\ref{tab:platforms} summarizes key
statistics. API endpoints, the collection funnel, filtering parameters, and topic details are in Appendix~\ref{app:dataset}, and ethics and licensing are in Appendix~\ref{app:ethics}. The resulting labels are deterministic reconstructions from public records, but
they remain behavioral proxies. \(\mathsf{G}_1\) uses on-chain and API position records to assign a binary label indicating whether the commenter held a financial stake at comment time. This is the most directly verifiable label, as it requires no inference beyond record presence. \(\mathsf{G}_2\) uses revealed position side as the target, not an unobserved mental state. \(\mathsf{G}_3\) depends on linking comments to subsequent position changes, which is cleaner on Manifold than on Polymarket. \(\mathsf{G}_4\) evaluates incremental collective odds-direction signal above a commitment-weighted baseline, not a causal effect of comments on market odds.
\begin{table}[tb]
\centering
\caption{Platform statistics for the two prediction-market sources used in StakeBench.}
\label{tab:platforms}
\scriptsize
\setlength{\tabcolsep}{3.2pt}
\renewcommand{\arraystretch}{1.18}
\resizebox{\linewidth}{!}{%
\begin{tabular}{@{}lrrrrcrrrrrl@{}}
\toprule
\rowcolor{tablehead}
& \multicolumn{4}{c}{\textbf{Corpus}} &
\multicolumn{2}{c}{\textbf{Comment activity}} &
\multicolumn{3}{c}{\textbf{Commitment labels}} &
\multicolumn{2}{c}{\textbf{Market-level signal}} \\
\cmidrule(lr){2-5}
\cmidrule(lr){6-7}
\cmidrule(lr){8-10}
\cmidrule(l){11-12}
\rowcolor{tablesubhead}
\textbf{Platform} &
\makecell{\textbf{Cleaned}\\\textbf{comments}} &
\makecell{\textbf{Markets}} &
\makecell{\textbf{Users}} &
\makecell{\textbf{Date}\\\textbf{range}} &
\makecell{\textbf{Median}\\\textbf{length}} &
\makecell{\textbf{Median}\\\textbf{comments}\\\textbf{/ market}} &
\makecell{\textbf{Position}\\\textbf{coverage}} &
\makecell{\(\boldsymbol{\mathsf{G}_1/\mathsf{G}_2}\)\\\textbf{instances}} &
\makecell{\(\boldsymbol{\mathsf{G}_3}\)\\\textbf{instances}} &
\makecell{\(\boldsymbol{\mathsf{G}_4}\)\\\textbf{windows}} &
\makecell{\textbf{Resolution}\\\textbf{source}} \\
\midrule
\textbf{Polymarket} &
514,314 & 2,070 & 24,909 &
\makecell{May 2024--\\Mar 2026} &
44 & 70 & 27.2\% & 139,848 & 73,893 & 603 &
\makecell[l]{Price-inferred\\\(\geq 0.90\)} \\
\addlinespace[2pt]
\textbf{Manifold} &
46,562 & 191 & 3,250 &
\makecell{Jan 2022--\\Mar 2026} &
107 & 163 & 80.5\% & 37,430 & 27,116 & 789 &
Direct API field \\
\bottomrule
\end{tabular}
}
\end{table}

\noindent
\textbf{Coverage and cleaning.}
Position side is reconstructed by replaying each commenter's public activity records (Polymarket) or bet history (Manifold) up to the comment time (Appendix~\ref{app:position-recon}). A comment whose author has no reconstructed position in the market receives no \(\mathsf{G}_2\) or \(\mathsf{G}_3\) label, and position coverage is 27.2\% on Polymarket against 80.5\% on Manifold. Table~\ref{tab:platforms} gives the instance counts after cleaning (Appendix~\ref{app:filters}): 73,893 \(\mathsf{G}_3\) instances from 139,848 \(\mathsf{G}_1\)/\(\mathsf{G}_2\) instances on Polymarket and 27,116 from 37,430 on Manifold. Comments removed at any step remain in the released files, and Section~\ref{sec:robustness} tests whether this reduction accounts for the \(\mathsf{G}_3\) result.

\subsection{Notation}
\label{sec:notation}

Table~\ref{tab:tasks} summarizes the variables used by the four tasks.
We use \(i\) for a comment instance. \(q_i\) is the corresponding market
question and \(x_i\) is the comment text. For \(\mathsf{G}_1\),
\(y_i\in\{0,1\}\) indicates whether the comment author has any reconstructed
position at comment time, with \(1\) mapped to YES and \(0\) to NO. For
\(\mathsf{G}_2\), \(z_i\in\{\mathrm{YES},\mathrm{NO}\}\) is the revealed side of
that position for positioned comments. \(s_i\geq0\) is the reconstructed stake
magnitude of the position, measured in USD for Polymarket and platform Mana for
Manifold; it is supplied to \(\mathsf{G}_3\) and used by Commitment-Calibrated Accuracy (CCA), Costly Signal Delta (CSD), and the
\(\mathsf{G}_4\) stake-weighted baseline. For \(\mathsf{G}_3\),
\(a_i\in\{\mathrm{flip},\mathrm{increase},\mathrm{decrease},\mathrm{hold}\}\)
is the first net position action observed within the 48-hour post-comment
window: \(\mathrm{flip}\) switches to the opposite side, \(\mathrm{increase}\)
grows exposure on the current side, \(\mathrm{decrease}\) shrinks exposure on
the current side, and \(\mathrm{hold}\) means no qualifying update is observed. For \(\mathsf{G}_4\), \(w\in\mathcal{W}\) indexes an evaluated aggregation
window for a single market question after the horizon and near-tie filters.
\(\mathcal{I}_w\) is the set of positioned comment instances in that window.
\(p_w^0\) is the mean YES price for the same market during the comment window;
the superscript \(0\) marks this pre-horizon reference window. \(h\) denotes the
prediction horizon, and \(p_w^h\) is the mean YES price in the subsequent price
window at horizon \(h\). The target direction is
\(d_w=\operatorname{dir}(p_w^h-p_w^0)\in\{\mathrm{UP},\mathrm{DOWN}\}\), with
near-zero moves filtered out as described in Section~\ref{sec:setup}. Let
\(b(\mathrm{YES})=+1\) and \(b(\mathrm{NO})=-1\). The stake-weighted baseline
direction is
\(\tilde{d}_w=\operatorname{dir}\!\left(\sum_{i\in\mathcal{I}_w}s_i b(z_i)\right)\),
and \(\tilde{d}=\{\tilde{d}_w\}_{w\in\mathcal{W}}\) denotes the corresponding
baseline sequence.


Model predictions carry hats (\(\hat{y}_i\), \(\hat{z}_i\), \(\hat{a}_i\),
\(\hat{d}_w\)), \(\hat{d}=\{\hat{d}_w\}_{w\in\mathcal{W}}\) is the
\(\mathsf{G}_4\) prediction sequence, and \(\bot\) denotes the NEUTRAL output
where abstention is allowed. Accuracy variants form one family
\(\mathrm{Acc}_{x}\) whose subscript names the scored subset:
\(\mathrm{Acc}_{\mathrm{strict}}\) counts \(\mathsf{G}_2\) abstentions as
errors, \(\mathrm{Acc}_{\mathrm{act}}\) is four-class \(\mathsf{G}_3\)
accuracy, and \(\mathrm{Acc}_f\) and \(\mathrm{Acc}_d\) score the flip-vs-hold
and decrease-vs-hold slices. Directed Accuracy (DA) scores only non-abstaining
\(\mathsf{G}_2\) predictions. CCA, CSD, CG and MCG are defined in
Section~\ref{sec:metrics}. PLA is \(\mathsf{G}_4\) direction accuracy,
\(\Delta_{\mathrm{PLA}}\) its advantage over the stake-weighted odds-direction
baseline \(\tilde{d}\), and \(n_+\) the number of splits with
\(\Delta_{\mathrm{PLA}}>0\). Agg is the cross-task score of
Appendix~\ref{app:detailed settings}. In the protocol column of
Table~\ref{tab:tasks}, ``binomial sig.'' is a one-sided test against the
task's chance level and ``descriptive'' means that no per-split threshold is
applied.

\begin{table}[tb]
\centering
\caption{Task definitions and metric families.}
\vspace{0mm}
\label{tab:tasks}
\scriptsize
\setlength{\tabcolsep}{1.35pt}
\renewcommand{\arraystretch}{1.18}
\begin{tabularx}{\linewidth}{@{}%
>{\centering\arraybackslash}p{0.042\linewidth}%
>{\centering\arraybackslash}p{0.140\linewidth}%
>{\centering\arraybackslash}p{0.095\linewidth}%
>{\centering\arraybackslash}p{0.078\linewidth}%
>{\centering\arraybackslash}p{0.055\linewidth}%
>{\centering\arraybackslash}X%
>{\centering\arraybackslash}p{0.142\linewidth}%
>{\centering\arraybackslash}p{0.145\linewidth}@{}}
\toprule
\rowcolor{tablehead}
\multicolumn{6}{c}{\textbf{Task setup}} &
\multicolumn{2}{c}{\textbf{Metrics}} \\
\cmidrule(lr){1-6}\cmidrule(l){7-8}
\rowcolor{tablesubhead}
\textbf{Task} & \textbf{Specific task} & \textbf{Input} &
\textbf{Prediction} & \textbf{Target} & \textbf{Target resolution} &
\makecell[c]{\textbf{Direct score}} &
\textbf{Protocol} \\
\midrule
\rowcolor{tablesubhead}
\(\mathsf{G}_1\)
& \makecell[c]{Who Holds\\a Position?}
& \(q_i,x_i\)
& \(\hat{y}_i\)
& \(y_i\)
& \makecell[c]{0/NO: no position\\1/YES: position exists}
& Acc, F1
& binomial sig. \\
\addlinespace[1pt]
\rowcolor{tablesubhead}
\(\mathsf{G}_2\)
& \makecell[c]{Which Side\\Is Revealed?}
& \(q_i,x_i\)
& \(\hat{z}_i\)
& \(z_i\)
& \makecell[c]{YES/NO: target side\\NEUTRAL: abstain output}
& \makecell[c]{\(\mathrm{Acc}_{\mathrm{strict}}\), DA\\CCA, CSD}
& \makecell[c]{binomial sig.\\CG/MCG} \\
\addlinespace[1pt]
\rowcolor{tablesubhead}
\(\mathsf{G}_3\)
& \makecell[c]{What Action\\Follows?}
& \(q_i,x_i,z_i,s_i\)
& \(\hat{a}_i\)
& \(a_i\)
& \makecell[c]{flip, increase,\\decrease, hold\\post-comment action}
& \makecell[c]{\(\mathrm{Acc}_{\mathrm{act}}\),\\\(\mathrm{Acc}_f\), \(\mathrm{Acc}_d\)}
& \makecell[c]{binomial sig.\\CG/MCG} \\
\addlinespace[1pt]
\rowcolor{tablesubhead}
\(\mathsf{G}_4\)
& \makecell[c]{Where Do\\ Aggregate\\Odds Move?}
& \(p_w^0,\mathcal{I}_w\)
& \(\hat{d}_w\)
& \(d_w\)
& \makecell[c]{UP/DOWN: target move\\NEUTRAL: abstain output\\near-ties filtered}
& \makecell[c]{PLA,\\\(\Delta_{\mathrm{PLA}}\), \(n_+\)}
& \makecell[c]{baseline \(\tilde{d}\)\\CG/MCG\\descriptive} \\
\bottomrule
\end{tabularx}
\end{table}

\subsection{Task Definitions}
\label{sec:tasks}

\textbf{\(\mathsf{G}_1\): Market Commitment Detection.}
This task tests whether a model can recognize financially committed speech
before downstream inference conditions on position.

\textbf{\(\mathsf{G}_2\): Revealed-Side Identification.}
This task tests whether a model can infer the side supported by a positioned
commenter's revealed market commitment.
Revealed side is a behavioral proxy, not a direct observation of private belief.
Strict accuracy counts abstention as incorrect on position-labeled examples,
while DA conditions on non-abstaining predictions.
Here \(\mathbf{1}[\cdot]\) denotes the indicator function: $\mathrm{DA} =
\frac{\sum_{i:\hat{z}_i\neq \bot}
\mathbf{1}[\hat{z}_i=z_i]}
{\left|\{i:\hat{z}_i\neq \bot\}\right|}.$
If a model produces no directional prediction for a split, DA is undefined for
that split and the split is excluded from the corresponding macro average.

\textbf{\(\mathsf{G}_3\): Future Action Anticipation.}
This task tests whether a model can anticipate how a positioned commenter will
change their exposure after speaking. It uses the action labels defined in
Section~\ref{sec:notation}. Thus \(\mathsf{G}_3\) uses a per-comment action
window, while \(\mathsf{G}_4\) uses market-level aggregation and price windows.
\(\mathrm{Acc}_{\mathrm{act}}\) is the four-class action accuracy. The two
binary slices are \(\mathrm{Acc}_f\) for flip-vs-hold and
\(\mathrm{Acc}_d\) for decrease-vs-hold; each slice is evaluated on examples
whose true action belongs to the named pair.

\textbf{\(\mathsf{G}_4\): Collective Odds Projection.}
This task tests whether aggregated comment signals add odds-direction signal
beyond the revealed position distribution. Using the window notation from
Section~\ref{sec:notation}, we score the target odds move against the model's
predicted direction. The baseline asks whether a model adds signal beyond the
revealed position distribution already present in the market.
We use PLA and its advantage over the baseline as
the evaluation metrics: $\mathrm{PLA}(\hat{d})=\frac{1}{|\mathcal{W}|}
\sum_{w\in\mathcal{W}}\mathbf{1}[\hat{d}_w{=}d_w],\Delta_{\mathrm{PLA}}=\mathrm{PLA}(\hat{d})-\mathrm{PLA}(\tilde{d}).$
\section{Evaluation Protocol}
\label{sec:eval-protocol}

StakeBench defines a three-layer evaluation protocol: (i)~a progressive task
suite linking language to position, action, and odds, (ii)~commitment-aware
metrics separating surface sentiment from revealed position side, and
(iii)~validity audits testing the assumption that financial commitment changes
language in measurable ways. All targets are derived from public records without
human annotation. \textbf{Supported claims.}
StakeBench supports claims about commitment-based supervision, relative task
difficulty, and transfer across platform incentive regimes. It does not by
itself identify causal effects of comments on market odds or fully separate model
failure from market efficiency. \textbf{Reference points.}
StakeBench uses null baselines for classification tasks, revealed-preference
reference scores for commitment gaps, and the commitment-weighted
odds-direction baseline for \(\mathsf{G}_4\). These reference points calibrate
the strength of the observed signal and prevent small score differences from
being overread as model orderings.

\subsection{Commitment-Aware Metrics}
\label{sec:metrics}

\textbf{CCA and CSD.}
For a model \(M\), CCA is the stake-weighted
version of DA for \(\mathsf{G}_2\), using \(\log(1+s_i)\) to limit the
influence of extreme positions. CSD compares DA between
the highest and lowest stake quartiles.
Let $\mathrm{DA}^{(k)}$ denote DA in the $k$-th stake-magnitude quartile
($k{=}1$ lowest stake, $k{=}4$ highest stake).
If the set \(\{i:\hat{z}_i\neq\bot\}\) is empty or the CCA denominator is zero, CCA is undefined for that split and excluded from macro averaging. Positive CSD is the costly-signaling prediction for committed speech.
\begin{equation}
\label{eq:commitment-weighted-metrics}
\begin{aligned}
\mathrm{CCA}(M)
&=
\frac{\sum_{i:\hat{z}_i\neq\bot}
  \log(1+s_i)\,\mathbf{1}[\hat{z}_i{=}z_i]}
{\sum_{i:\hat{z}_i\neq\bot}
  \log(1+s_i)}, \quad \mathrm{CSD}(M)=
\mathrm{DA}^{(4)}-\mathrm{DA}^{(1)}.
\end{aligned}
\end{equation}

\textbf{CG and MCG.}
For a model \(M\), let \(O_j\) be the task-specific revealed-preference
reference score: the score obtained from the observed market commitment
baseline for task \(j\), rather than from a text-only model prediction. Concretely, \(O_2\) uses the stake-weighted market-majority reference,
\(O_3\) uses the realized post-comment action, and \(O_4\) uses the
stake-weighted odds-direction baseline \(\tilde{d}\).
\(S(M,j)\) is corresponding model score for task \(j\in\{2,3,4\}\)
(\(S(M,2){=}\mathrm{DA}\), \(S(M,3){=}\mathrm{Acc}_{\mathrm{act}}\),
\(S(M,4){=}\mathrm{PLA}\), with \(\mathrm{PLA}\) evaluated on model \(M\)'s
\(\hat{d}\)). \(\mathsf{G}_1\) is excluded because it lacks a
revealed-preference reference score.
\begin{equation}
\label{eq:commitment-gap}
\mathrm{CG}(M,j)=O_j-S(M,j),
\qquad
\mathrm{MCG}(M)=\tfrac{1}{3}\sum_{j=2}^{4}\mathrm{CG}(M,j).
\end{equation}
\(\mathrm{CG}\) is signed: positive values indicate a shortfall from the
revealed-preference reference score, negative values indicate that the model
exceeds it, and values closer to zero indicate closer alignment.
For \(\mathsf{G}_2\), CG compares analogous but not identical targets; Appendix~\ref{app:theory}
defines the reference scores.

\subsection{Commitment Validity Audits}

\(\mathsf{G}_2\) performance might reflect smart-money effects, self-selection, or outcome
prediction rather than language shaped by commitment. We therefore run three
falsification audits alongside the benchmark scores. These audits test specific
alternative explanations rather than certifying that every reconstructed label is noise-free. Full results in Appendix~\ref{app:audits}. Throughout, \(z_i\) is the commenter's true position side, \(\hat{z}_i\) the model's predicted side, and \(r_i\) the market's resolved side. \textbf{Wrong-Bettor Test.}
We isolate comments where $z_i\!\neq\!r_i$ (the trader held the losing side).
A model exploiting market truth predicts $r_i$. A model reading
commitment language predicts $z_i$. We test
$H_0\colon\Pr(\hat{z}_i{=}z_i){=}0.5$ (one-sided
binomial) on directional (non-abstaining) predictions only, requiring
$n_d{\geq}20$ per topic--platform split, where $n_d$ denotes
the number of directional predictions for the model--split pair. \textbf{Position-Size Dose Response.}
We partition directional \(\mathsf{G}_2\) examples into stake-magnitude quartiles
and test whether CSD is positive via one-sided Mann--Whitney
with $n_d{\geq}40$ directional predictions and non-constant stake magnitudes.
Here, $n_d$ is defined as above. \textbf{Within-User Difference-in-Differences (DiD).}
We compare the same user's comments before and after their first position in the same market.
The DiD statistic is defined as $\mathrm{DiD}=\mathrm{DA}^{+}_{r}-\mathrm{DA}^{-}_{r}$. It is post-position DA minus pre-position DA, both scored against resolution $r_i$, where superscripts \(+\) and \(-\) denote post-position and pre-position comments, and the subscript \(r\) indicates scoring against $r_i$.
We treat negative DiD (language drifting away from market truth after position-taking) as a signature of commitment bias.


\section{Experimental Setup}
\label{sec:setup}

The benchmark evaluation covers 15 large language models spanning three categories: three closed-source models include GPT-5.5~\citep{openai2025gpt55}, Gemini-2.5-Flash~\citep{comanici2025gemini}, and Claude-Haiku-4.5~\citep{anthropic2025haiku45}; eight general-purpose open-weight models include Qwen3-32B, Qwen3-14B, Qwen3-30B-A3B, and Qwen3-8B~\citep{yang2025qwen3}, Gemma2-9B~\citep{team2024gemma}, DeepSeek-R1-Distill-Llama-8B~\citep{guo2025deepseek}, Llama-3.2-3B-Instruct~\citep{grattafiori2024llama}, and Mistral-7B-Instruct-v0.3~\citep{mistral7b}; The four finance-domain models include Fino1-14B (based on Qwen2.5-14B) and Fino1-8B (based on Llama-3.1-8B)~\citep{qian2025fino1}, Finance-Chat-7B (based on Llama-2-7B-Chat)~\citep{cheng2024finchat}, and FinMA-30B (based on Llama-2-30B)~\citep{xie2023pixiu}. All models are evaluated under a fixed prompting protocol with greedy decoding (temperature \(0\), seed \(42\)). Macro scores average uniformly over the 18 topic--platform splits rather than pooling comments, preventing high-volume markets from dominating the benchmark. \(\mathsf{G}_4\) uses adaptive horizons in the 1--30 day range and removes near-tie odds moves with \(|p_w^h-p_w^0|\leq0.005\). Classification tasks are evaluated with one-sided binomial tests against task-specific chance baselines, while \(\mathsf{G}_4\) is reported descriptively relative to the stake-weighted odds-direction baseline. Other settings are in Appendix~\ref{app:detailed settings}.

\section{Results}
\label{sec:results}

\begin{table}[tb]
\centering
\caption{Main benchmark results with macro-averaged scores. Red/blue mark the
best/second-best model values.}
\label{tab:main}
\tiny
\setlength{\tabcolsep}{2pt}
\renewcommand{\arraystretch}{1.04}
\resizebox{\linewidth}{!}{%
\begin{tabular}{lrrrrrrrrrrrrr}
\toprule
\rowcolor{tablehead}
\textbf{Model} & \multicolumn{2}{c}{\(\boldsymbol{\mathsf{G}_1}\)}
& \multicolumn{4}{c}{\(\boldsymbol{\mathsf{G}_2}\)}
& \multicolumn{3}{c}{\(\boldsymbol{\mathsf{G}_3}\)}
& \multicolumn{3}{c}{\(\boldsymbol{\mathsf{G}_4}\)}
& \textbf{Agg} \\
\cmidrule(lr){2-3}\cmidrule(lr){4-7}\cmidrule(lr){8-10}\cmidrule(lr){11-13}
\rowcolor{tablesubhead}
& \textbf{Acc} & \textbf{F1}
& \(\boldsymbol{\mathrm{Acc}_{\mathrm{strict}}}\) & \textbf{DA} & \textbf{CCA} & \textbf{CSD}
& \(\boldsymbol{\mathrm{Acc}_{\mathrm{act}}}\) & \(\boldsymbol{\mathrm{Acc}_f}\) & \(\boldsymbol{\mathrm{Acc}_d}\)
& \textbf{PLA} & $\boldsymbol{\Delta}_{\mathrm{PLA}}$
& $\boldsymbol{n_+}$ & \\
\midrule
\rowcolor{tablesubhead}\multicolumn{14}{@{}l}{\textit{Closed-source LLMs}} \\
\quad GPT-5.5          & .689 & .598 & .448 & \best{.599} & \best{.678} & .048 & .279 & \second{.564} & .514 & \second{.521} & \second{-.052} & \best{5} & \best{.204} \\
\quad Gemini-2.5-Flash & .711 & .666 & .384 & .587 & .623 & -.056 & .278 & .538 & .500 & .392 & -.181 & 3 & .175 \\
\quad Claude-Haiku-4.5 & .697 & .616 & .381 & \second{.594} & \second{.655} & .057 & .257 & .522 & .510 & .363 & -.210 & 1 & .138 \\
\addlinespace[2pt]
\rowcolor{tablesubhead}\multicolumn{14}{@{}l}{\textit{General open-weight LLMs}} \\
\quad Qwen3-32B        & \best{.747} & \best{.711} & .430 & .578 & .606 & .030 & .256 & .481 & \second{.524} & .402 & -.165 & \second{4} & \second{.176} \\
\quad Qwen3-30B        & .670 & .562 & .344 & .548 & .609 & \second{.092} & \second{.294} & \best{.575} & .512 & .209 & -.373 & 1 & .140 \\
\quad Qwen3-14B        & \second{.740} & \second{.703} & .423 & .567 & .621 & .022 & .274 & .554 & .493 & .345 & -.222 & 2 & .173 \\
\quad Gemma2-9B        & .718 & .673 & .397 & .578 & .622 & -.038 & .166 & .381 & .513 & .310 & -.257 & 3 & .152 \\
\quad DeepSeek-R1-8B   & .721 & .696 & .345 & .514 & .534 & .054 & .230 & .434 & .418 & .326 & -.256 & 2 & .116 \\
\quad Qwen3-8B         & .692 & .612 & .325 & .564 & .607 & .038 & .168 & .474 & .377 & .186 & -.381 & 0 & .103 \\
\quad Mistral-7B       & .534 & .169 & .335 & .523 & .552 & -.050 & .169 & .382 & .397 & .267 & -.300 & 2 & .045 \\
\quad Llama-3.2-3B & .567 & .322 & \best{.534} & .534 & .535 & -.008 & \best{.401} & .395 & \best{.665} & .159 & -.408 & 1 & .092 \\
\addlinespace[2pt]
\rowcolor{tablesubhead}\multicolumn{14}{@{}l}{\textit{Finance-domain open-weight LLMs}} \\
\quad FinMA-30B       & .503 & .668 & .476 & .506 & .519 & .031 & .164 & .342 & .413 & .130 & -.437 & 0 & .003 \\
\quad Fino1-14B       & .648 & .498 & .363 & .565 & .608 & .078 & .190 & .457 & .415 & .275 & -.307 & 1 & .096 \\
\quad Fino1-8B        & .601 & .414 & \second{.524} & .524 & .573 & \best{.094} & .162 & .366 & .404 & \best{.560} & \best{-.022} & 2 & .072 \\
\quad Fin-Chat-7B     & .531 & .169 & .445 & .514 & .552 & .003 & .167 & .472 & .519 & .433 & -.134 & \best{5} & .073 \\
\bottomrule
\end{tabular}
}
\end{table}

Results are organized around the claims StakeBench can support.
We first report the full benchmark scores, then analyze the position-side, action, and odds layers that reveal where current models fail.
We emphasize robust patterns, class
collapse, transfer behavior, and baseline-relative signal rather than treating
small score differences as final model rankings.

Table~\ref{tab:main} reports
macro-averaged results for all 15 models.
Models cover all 18 topic--platform combinations unless explicitly noted.
The reference scores are \(O_2{=}0.552\) for \(\mathsf{G}_2\)
(stake-weighted market majority predicting resolution, see
Appendix~\ref{app:theory}), \(O_3{=}1.0\) for \(\mathsf{G}_3\), 
and $O_4$
Table~\ref{tab:main} reports \(\mathsf{G}_4\) through \(\mathrm{PLA}\), $\Delta_{\mathrm{PLA}}$ and \(n_+\);
the corresponding
\(O_4\) values are reported with commitment gaps in Table~\ref{tab:cg}.
MCG spans 0.242 (GPT-5.5, lowest MCG) to 0.440
(FinMA-30B, highest MCG), with mean MCG 0.278 for closed-source models and 0.360
for finance-domain models (Table~\ref{tab:cg}).
Finance-domain models trail on speaker-level tasks, and paired open-weight
comparisons show little benefit from finance tuning
(Table~\ref{tab:finance-pairs}).

\begin{figure}[tb]
    \centering
    \resizebox{\linewidth}{!}{%
        \includegraphics{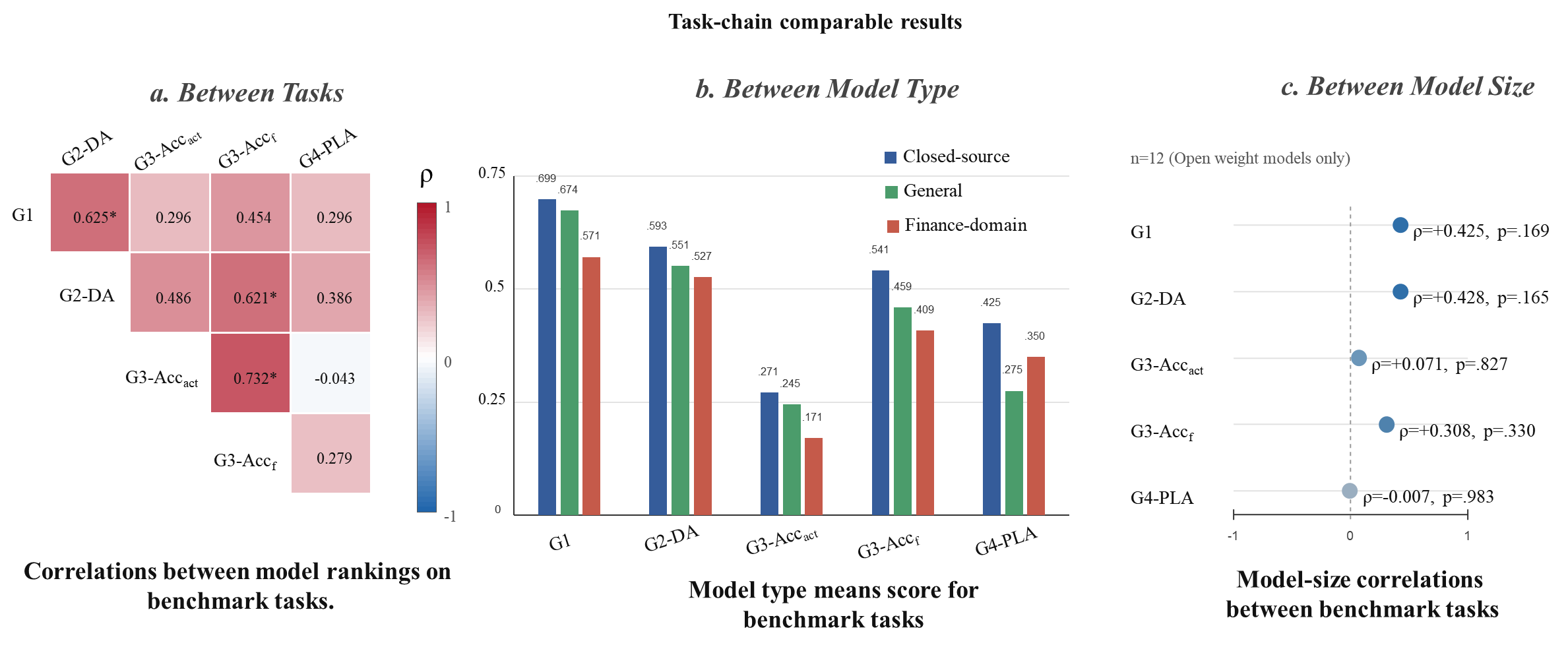}%
    }
    \caption{Task-chain correlations, model-family means, and scale effects.
    Asterisks mark \(p{<}0.05\).}
    \label{fig:task-chain-summary}
\end{figure}

\subsection{\(\mathsf{G}_1\): Market Commitment Detection}

Eleven of fifteen models exceed \(\mathsf{G}_1\) accuracy 0.65, with Qwen3-32B leading at
0.747 (F1\,0.711). FinMA-30B (0.503) and Mistral-7B (0.534) fall near chance,
although their F1 scores differ substantially.
The Polymarket--Manifold gap is negligible
(\(\delta{=}0.016\), \(p{=}0.73\); Appendix
Table~\ref{tab:aggregate-analysis}), and cross-platform rankings are nearly
identical (\(\rho{=}0.975\)).
This
indicates that committed speech carries recognizable discourse-level markers
that are stable across both platforms' writing styles.
This near-perfect rank transfer stands in sharp contrast to \(\mathsf{G}_2\) and \(\mathsf{G}_3\), where
platform effects are large, marking market commitment detection as the most
transferable layer of the task hierarchy.

\subsection{\(\mathsf{G}_2\): Revealed-Side Identification}
\label{sec:g2}

\(\mathsf{G}_2\) DA spans 0.506 to 0.599, with closed-source models averaging 0.593 and
finance-domain models 0.527. The ceiling near 0.60 indicates that many
position sides are not directly recoverable from surface sentiment.
GPT-5.5's CCA\,$=0.678$ exceeds its DA\,$=0.599$, showing stronger
signals in higher-stake comments. Finance-domain models instead show
outcome-oriented behavior: FinMA-30B scores 0.506 against commenter side but
0.538 against market resolution, while most general models show the opposite
pattern (Table~\ref{tab:outcome-side-audit}). \(\mathsf{G}_2\) is also platform-sensitive
($\rho{=}0.286$ across shared-topic platform rankings).
Together, these patterns suggest that StakeBench is not merely measuring
generic financial optimism. Models must distinguish the
speaker's committed side from the market outcome, and this distinction is where
finance-domain tuning appears least reliable.

\subsection{\(\mathsf{G}_3\): Future Action Anticipation}
\label{sec:g3}

\begin{table}[tbh]
\centering
\caption{Per-class recall profile for \(\mathsf{G}_3\). Model rows report
recall by true action class; the Ground truth row reports the true action
distribution.}
\label{tab:g3-action-profile}
\footnotesize
\setlength{\tabcolsep}{3pt}
\renewcommand{\arraystretch}{1.08}
\begin{tabular}{@{}llrrrr@{}}
\toprule
\rowcolor{tablehead}
\textbf{Type} & \textbf{Model} & \textbf{Flip\%} & \textbf{Increase\%} & \textbf{Decrease\%} & \textbf{Hold\%} \\
\midrule
\rowcolor{tablesubhead}
\multirow{5}{*}{\makecell[l]{Non-degenerate}}
& GPT-5.5          & 11.0 & 41.4 & 21.8 & 35.6 \\
& Gemini-2.5-Flash &  7.0 & 21.1 & 27.1 & 54.6 \\
& Claude-Haiku-4.5 &  7.6 & 27.1 & 34.4 & 32.5 \\
& Qwen3-32B        & 24.8 & 22.0 & 11.8 & 36.1 \\
& Qwen3-14B        &  8.4 &  9.0 & 19.9 & 47.5 \\
\addlinespace[2pt]
\rowcolor{tablesubhead}
\multirow{3}{*}{\makecell[l]{Partially degenerate}}
& Qwen3-30B      & 23.9 &  3.4 & 28.9 & 42.1 \\
& DeepSeek-R1-8B & 60.5 & 11.5 & 20.6 & 15.8 \\
& Llama-3.2-3B   & 26.9 &  4.6 &  0.0 & 70.8 \\
\addlinespace[2pt]
\rowcolor{tablesubhead}
\multirow{7}{*}{Degenerate}
& FinMA-30B       &  1.1 &  0.0 & 97.0 &  0.0 \\
& Fino1-14B       &  5.8 &  4.9 & 84.4 &  5.3 \\
& Gemma2-9B       &  7.1 &  0.3 & 92.0 &  0.1 \\
& Qwen3-8B        & 44.3 &  3.2 & 42.7 &  3.3 \\
& Fino1-8B        & 79.1 &  0.0 & 18.4 &  0.0 \\
& Mistral-7B      & 90.2 &  0.0 & 11.2 &  0.0 \\
& Fin-Chat-7B     & 100.0  &  0.0 &  0.0 &  0.0 \\
\midrule
\rowcolor{tablesubhead}
\multicolumn{2}{@{}l}{Cross-model mean} & 33.2 &  9.9 & 34.0 & 22.9 \\
\multicolumn{2}{@{}l}{Ground truth}  &  6.0 & 19.9 & 26.5 & 47.6 \\
\bottomrule
\end{tabular}
\end{table}

\(\mathsf{G}_3\) exposes a class-coverage collapse that aggregate scores
conceal. Ten of fifteen models show usable recall on only one or two action
classes: some models favor
\textit{flip}, while others favor \textit{decrease}
(Table~\ref{tab:g3-action-profile}). Mean \textit{hold} recall is 22.9\%,
even though \textit{hold} is the largest true class (47.6\%), suggesting that
models over-predict behavioral change. \(\mathsf{G}_3\) is easier on Manifold
(\(\delta{=}-0.076\) for \(\mathrm{Acc}_{\mathrm{act}}\) and
\(\delta{=}-0.129\) for \(\mathrm{Acc}_f\)). Part of this gap may reflect
label coverage, which differs between platforms: 73,893 of the 139,848
\(\mathsf{G}_1\)/\(\mathsf{G}_2\) instances on Polymarket carry a
\(\mathsf{G}_3\) label, against 27,116 of 37,430 on Manifold
(Section~\ref{sec:dataset}). Whether the unreconstructed Polymarket records
differ systematically from the retained ones cannot be tested from the
released data. On Manifold, where the full trading distribution is
observable, restricting evaluation to the most active commenters leaves
macro-recall unchanged (Section~\ref{sec:robustness}), so activity skew does
not account for the collapse where it can be tested. The platform comparison
is in Appendix Table~\ref{tab:aggregate-analysis}. Scores are stable across 6h--72h action windows
(Table~\ref{tab:g3-window-sensitivity}), so we retain 48h as the default horizon.
This failure mode is central to StakeBench. A model can identify a position or
even infer its side while still failing to reason about whether the speaker will
flip, increase, decrease, or hold that exposure. The gap between language
interpretation and later action is therefore not a small degradation, but a
change in the kind of reasoning required.

\subsection{\(\mathsf{G}_4\): Collective Odds Projection}
\label{sec:g4}

No model consistently exceeds the stake-weighted baseline $\tilde{d}$ in aggregate.
All $\Delta_{\mathrm{PLA}}$ values are negative ($-0.022$ to $-0.437$), showing
that LLMs do not add collective odds-direction signal beyond the
commitment-weighted majority. This is a baseline-relative result, not evidence
that comments lack market-odds information. Positive gains concentrate on Manifold
(27 of 32 model--split instances, Table~\ref{tab:g4-platform-gains}), consistent
with more residual signal on the play-money platform. \(\mathsf{G}_4\) is uncorrelated with \(\mathsf{G}_1\)--\(\mathsf{G}_3\) ($|\rho|{\leq}0.39$),
showing that speaker-level tasks and crowd-level odds projection measure different
abilities.
Figure~\ref{fig:task-chain-summary}(a) places this separation in the task chain:
\(\mathsf{G}_1\)--\(\mathsf{G}_2\) ($\rho{=}0.625$, $p{=}0.013$) and
\(\mathsf{G}_2\)--\(\mathsf{G}_3\)-\(\mathrm{Acc}_f\)
($\rho{=}0.621$, $p{=}0.013$) are significant, but the chain breaks at
\(\mathsf{G}_4\). Figure~\ref{fig:task-chain-summary}(c) further shows that model size is not
significantly correlated with any task score across the 3B--32B open-weight
range; we report \(|\rho|\) because the largest absolute association is small
(\(|\rho|{\leq}0.43\), \(p{>}0.16\) across 12 models). The \(\mathsf{G}_4\) result is therefore a scope boundary on what the
benchmark can claim rather than a verdict on models. The benchmark does not
ask whether comments are correlated with future odds in isolation. It asks
whether a language model can add value after the market's revealed position
distribution has already been observed, and Section~\ref{sec:robustness}
tests whether window text carries any such residual signal at all.

\subsection{Data Limits and Model Limits}
\label{sec:robustness}

The \(\mathsf{G}_3\) and \(\mathsf{G}_4\) failures admit two readings: the
models cannot read a signal that the data carry, or the data carry little
signal to read. The analyses below separate these readings where the data allow (full tables and boundaries in Appendix~\ref{app:robust}).

\textbf{Lexical recoverability.}
A TF-IDF logistic-regression probe trained on comment text alone, with every
training comment preceding every test comment, reproduces the difficulty
ordering of the task hierarchy: balanced accuracy 0.685 and 0.729 for
\(\mathsf{G}_1\) (Manifold and Polymarket, floor 0.500), 0.531 and 0.521 for
\(\mathsf{G}_2\), and 0.246 and 0.228 for \(\mathsf{G}_3\) (floor 0.250),
with the same ordering under market-disjoint and commenter-disjoint splits.
The \(\mathsf{G}_1\) margin partly re-measures the explicit position-language
rule that selects the positive pool (Appendix~\ref{app:markers}), so it shows
that explicit commitment language is reliably detectable, not that latent
commitment is recoverable from arbitrary text. The \(\mathsf{G}_3\) floor
carries no such caveat: the action collapse of Section~\ref{sec:g3} is not a
failure to exploit a shallow cue that sits in the text, although a linear
probe bounds only shallow lexical recoverability. On the audit side, five of
the 25 Wrong-Bettor cells with uncorrected \(p<0.05\) survive
Benjamini--Hochberg correction (\(q<0.05\), peak follow-rate 0.844), four of
them in one topic, so commitment-following is detectable where present but
is not a uniform property of the discourse.

\textbf{Activity skew.}
If the \(\mathsf{G}_3\) subset over-represents active traders, the collapse
could be a sampling artifact. Manifold's 80.5\% position coverage makes this
testable there. Keeping only the 14.2\% most active commenters changes
macro-recall by \(-0.005\) (95\% CI \([-0.014,\,+0.004]\)) and raises
collapse by \(+0.018\) (\([+0.008,\,+0.033]\)), and macro-recall varies by at
most 0.021 across commenter-activity quartiles. Collapse moves in the
hypothesized direction, but by an order of magnitude less than the gap
between mean \textit{hold} recall (22.9\%) and the \textit{hold} base rate
(47.6\%), so selection on activity does not account for it. This is a
Manifold stress test under one activity skew, not proof that all forms of
Polymarket selection bias are eliminated.

\textbf{Residual odds signal.}
The stake-weighted baseline \(\tilde{d}\) absorbs the revealed positions in a
window, so \(\mathsf{G}_4\) scores only what text adds beyond them. A
supervised probe given the full window text scores, on held-out windows,
0.513 against a 0.547 stake-weighted baseline on Manifold (1,163-window pool)
and 0.643 against a 0.631 majority baseline on Polymarket (557), with paired
differences against the stronger baseline of \(-0.034\)
\([-0.142,\,+0.071]\) and \(+0.012\) \([-0.066,\,+0.084]\) (market-clustered
bootstrap). The negative
\(\Delta_{\mathrm{PLA}}\) of every model in Table~\ref{tab:main} is thus
consistent with text adding no direction signal at the window level, not
only with models failing to read it. The probe windows are a separate
reconstruction rather than the inputs the models saw, and no efficiency
quantity is measured, so this supports the efficiency reading of
\(\mathsf{G}_4\) without establishing it.

\textbf{Protocol variants.}
Two protocol changes already reported in the paper leave the
\(\mathsf{G}_3\) result in place. The binary slices of Table~\ref{tab:main}
average 0.462 (flip vs.\ hold) and 0.478 (decrease vs.\ hold) over the 15
models against a uniform-random reference of 0.500 (the binary subsets are
not balanced, so 0.500 is not empirical chance), and the 6h--72h horizon
sweep of Table~\ref{tab:g3-window-sensitivity} moves mean accuracy by less
than one percentage point. Both change label space and protocol rather than
wording~\citep{sclar2024quantifying}. Prompt templates are in
Appendix~\ref{app:prompts}.

\begingroup
\newcommand{\cYes}{\textcolor{green!45!black}{\textbf{YES}}}
\newcommand{\cNo}{\textcolor{red!70!black}{\textbf{NO}}}
\newcommand{\cNeu}{\textcolor{gray}{\textbf{NEU}}}
\newcommand{\cFlip}{\textcolor{magenta!70!black}{\textbf{flip}}}
\newcommand{\cHold}{\textcolor{gray}{\textbf{hold}}}
\newcommand{\cInc}{\textcolor{yellow!50!red}{\textbf{inc.}}}
\newcommand{\cDec}{\textcolor{red!70!black}{\textbf{dec.}}}
\newcommand{\cUp}{\textcolor{yellow!50!red}{\textbf{UP}}}
\newcommand{\cDown}{\textcolor{blue!70!black}{\textbf{DOWN}}}

\begin{table}[tb]
\centering
\tiny
\setlength{\tabcolsep}{1.1pt}
\renewcommand{\arraystretch}{1.16}
\caption{%
Representative case studies for \(\mathsf{G}_1\)--\(\mathsf{G}_4\).}
\label{tab:case-studies}
\begin{tabularx}{\linewidth}{@{}%
>{\centering\arraybackslash}p{0.035\linewidth}%
>{\raggedright\arraybackslash\hspace{0pt}}p{0.125\linewidth}%
>{\raggedright\arraybackslash\hspace{0pt}}X%
>{\raggedright\arraybackslash\hspace{0pt}}p{0.130\linewidth}%
>{\raggedright\arraybackslash\hspace{0pt}}p{0.130\linewidth}%
>{\raggedright\arraybackslash\hspace{0pt}}p{0.125\linewidth}%
>{\raggedright\arraybackslash\hspace{0pt}}p{0.155\linewidth}@{}}
\toprule
\rowcolor{tablehead}
\textbf{Task} &
\textbf{Market} &
\textbf{Comment cue} &
\textbf{Prediction} &
\textbf{Target resolution} &
\textbf{15-model summary} &
\textbf{Insight} \\
\midrule

\rowcolor{tablesubhead}
\(\mathsf{G}_1\) &
\textit{Russia \(\times\) Ukraine ceasefire in March?}\newline
\textcolor{gray}{Polymarket} &
\textit{``\textbf{I bought} with the rules \ldots{}
illegitimate and fraudulent.''} &
GPT, Gemini, Claude: \cYes\newline
Mistral, Qwen-30B: \cNo &
Position exists\newline
YES stake: \$159\newline
Target: \cYes / True &
8 \cYes{} / 7 \cNo\newline
Acc. 53.3\% &
First-person ownership is masked by complaint tone and third-person
``yes holders'' framing. \\

\addlinespace[2pt]

\rowcolor{tablesubhead}
\(\mathsf{G}_2\) &
\textit{US recession before end-2024?}\newline
\textcolor{gray}{Manifold} &
\textit{``10y2y treasury yield spread: it just
\textbf{disinverted}.''} &
GPT: \cYes\newline
Gemini, Claude, FinMA: \cNo &
Revealed side\newline
YES stake: \$1,364\newline
Target: \cYes &
7 \cYes{} / 7 \cNo\newline
1 \cNeu\newline
Acc. 46.7\% &
Correct reading needs macro timing knowledge, not only finance vocabulary. \\

\addlinespace[2pt]

\rowcolor{tablesubhead}
\(\mathsf{G}_3\) &
\textit{Will Manifold solve my puzzle? (Round 2)}\newline
\textcolor{gray}{Manifold} &
\textit{``I \textbf{spent 2500 mana on No} shares.''} &
Mistral, Qwen-30B: \cFlip\newline
GPT-5.5: \cInc\newline
Claude, Gemini: \cHold &
48h action\newline
NO \$2,501 \(\rightarrow\) YES\newline
Target: \cFlip &
5 \cFlip{} / 5 \cHold\newline
3 \cDec{} / 2 \cInc\newline
Acc. 33.3\% &
Past position disclosure does not reveal whether exposure will grow, shrink,
hold, or reverse. \\

\addlinespace[2pt]

\rowcolor{tablesubhead}
\(\mathsf{G}_4\) &
\textit{MicroStrategy BTC purchase Jul. 1--7?}\newline
\textcolor{gray}{Polymarket} &
\textit{``\textbf{100\% will be another buy}''};\,
\textit{``99\% for him to buy more''};\,
\textit{``buy more NO \ldots{} free money''}.\newline
\textit{``\textbf{not gonna smash ATM}''};\,
\textit{``pump and Dump by whales''}. &
DeepSeek, Claude, Gemini: \cDown\newline
GPT-5.5, FinMA: \cNeu\newline
Most others: \cUp &
YES odds\newline
0.810 \(\rightarrow\) 0.452\newline
stat\_3c: \cUp\newline
Target: \cDown &
3 \cDown{} / 9 \cUp\newline
3 \cNeu\newline
Acc. 20.0\% &
The crowd sounds bullish, but minority bearish cues better anticipate the next
odds move. \\

\bottomrule
\end{tabularx}
\end{table}
\endgroup

\subsection{Diagnostic Case Studies and Platform Effects}

Table~\ref{tab:case-studies} presents four representative task cases, and Appendix~\ref{app:examples} adds verified cases with per-model predictions. The examples show that discourse context can obscure commitment cues, domain timing can determine revealed side, position disclosure need not imply later action, and bullish aggregate language can precede a lower next-window price. These cases clarify the diagnostic role of StakeBench. The relevant signal is often not the surface polarity of the comment, but its relation to an observable commitment, later action, or price-window outcome. The same heterogeneity appears
at the model-family and platform levels. Model-family means are shown in
Figure~\ref{fig:task-chain-summary}(b), and cross-platform breakdowns appear in
Table~\ref{tab:aggregate-analysis} (Appendix~\ref{app:platform}).
Finance-domain models trail on speaker-level metrics (mean
\(\mathsf{G}_3\)-\(\mathrm{Acc}_{\mathrm{act}}\) 0.171 vs.\
0.271 for closed-source) despite a marginal \(\mathsf{G}_4\) PLA advantage, suggesting that
financial pretraining is associated with outcome-oriented rather than
position-side reasoning.
\(\mathsf{G}_1\) transfers fully across platforms ($\delta{=}0.016$, $\rho{=}0.975$).
\(\mathsf{G}_2\) aggregate performance is stable but model orderings diverge ($\rho{=}0.286$).
\(\mathsf{G}_3\) transfers at neither level ($\rho{\leq}0.207$, $p{\leq}0.001$).
\(\mathsf{G}_4\) model orderings are consistent ($\rho{=}0.771$).
This hierarchy (\(\mathsf{G}_1\) generalizes, \(\mathsf{G}_2\) partially,
\(\mathsf{G}_3\) not at all, \(\mathsf{G}_4\) distinct)
confirms that StakeBench's dual-platform design adds diagnostic coverage rather
than redundant replication.
\section{Conclusion}
\label{sec:conclusion}

We present StakeBench, an evaluation framework for commitment-grounded language
understanding that links 560,876 comments from 2,261 resolved markets across
Polymarket and Manifold to each author's verified position record, supporting
four progressive diagnostic tasks and three commitment-aware metrics grounded
in revealed preferences rather than perceived sentiment. Its contribution is an
auditable evaluation protocol with explicit behavioral targets, not a new model
or a final model ranking.
Our evaluation of 15 LLMs across 18 topic--platform combinations shows that
models partially recover individual commitment signals, with \(\mathsf{G}_2\) Directed
Accuracy spanning 0.506 to 0.599, but fail on future action and collective odds tasks.
\(\mathsf{G}_3\) collapses to one or two action labels in 10 of 15 models, and no
model consistently exceeds the stake-weighted majority baseline on \(\mathsf{G}_4\).
Financial pretraining is associated with weaker speaker-level reasoning than
comparable general models, model scale is uncorrelated with performance, and
most positive \(\mathsf{G}_4\) gains occur on the play-money platform.
Validity audits test whether committed speech differs from non-committed speech
in language~\citep{spence1978job,crawford1982} and whether models
exploit commitment cues or merely outcome-prediction cues.
These results should be read as evidence about observable commitment proxies
under the StakeBench protocol, not as claims about private beliefs or causal
market-odds formation.
Together, these findings show that commitment-grounded language understanding
remains an open challenge for LLMs across speaker positions,
behavioral consistency, and collective odds signals, suggesting that training
objectives using position context may be more productive than domain-specific
financial pretraining.

\smallskip\noindent\textbf{Limitations.}
CSD and DiD audits are underpowered on some topics. \(\mathsf{G}_3\) label coverage is lower on Polymarket than on Manifold. The benchmark uses fixed prompts and LLM probes rather than trainable reference models, so it measures behavior under that protocol rather than best achievable performance, and a prompt-wording ablation is left to future work. Public comments may enter future pretraining corpora, platform incentives differ, and \(\mathsf{G}_4\) may reflect market efficiency as well as model limits,
and all data are in English.

\smallskip\noindent\textbf{Ethics and broader impact.}
All data are sourced from public APIs. User identifiers are platform pseudonyms
(\texttt{proxy\_wallet} on Polymarket, \texttt{user\_id} on Manifold), and no
real names, IP addresses, or private messages are included.
Position-holding commenters differ from non-holders on observable dimensions
with effect sizes below $d{=}0.3$ (Cohen's~$d$), indicating modest but non-zero
selection bias in \(\mathsf{G}_2\) and \(\mathsf{G}_3\) evaluation sets.
The benchmark dataset, evaluation code, model outputs, and scripts are packaged
under CC-BY 4.0 to support review and reproduction.
The underlying platform data are used under Polymarket's Terms of Use and
Manifold's Terms of Service, both of which permit non-commercial research use.
Open-weight model weights follow their respective licenses (Meta Llama 3
Community License for Llama 3 and Apache 2.0 for Qwen3).

\begin{ack}
This work was funded by UKRI EPSRC (grant No. EP/Y028392/1): AI for Collective Intelligence (AI4CI). 
\end{ack}

\bibliographystyle{plainnat}
\bibliography{references}

@incollection{spence1978job,
  title={{Job market signaling}},
  author={Spence, Michael},
  booktitle={Uncertainty in Economics},
  pages={281--306},
  year={1978},
  publisher={Elsevier}
}

@article{crawford1982,
  title={{Strategic Information Transmission}},
  author={Crawford, Vincent P and Sobel, Joel},
  journal={Econometrica: Journal of the Econometric Society},
  pages={1431--1451},
  year={1982},
  publisher={JSTOR}
}

@article{samuelson1948,
  title={{Consumption theory in terms of revealed preference}},
  author={Samuelson, Paul A},
  journal={Economica},
  volume={15},
  number={60},
  pages={243--253},
  year={1948},
  publisher={JSTOR}
}

@article{halawi2024,
  title={{Approaching human-level forecasting with language models}},
  author={Halawi, Danny and Zhang, Fred and Yueh-Han, Chen and Steinhardt, Jacob},
  journal={Advances in Neural Information Processing Systems},
  volume={37},
  pages={50426--50468},
  year={2024}
}

@inproceedings{karger2025,
  title={{{ForecastBench}: A dynamic benchmark of {AI} forecasting capabilities}},
  author={Karger, Ezra and Bastani, Houtan and Yueh-Han, Chen and Jacobs, Zachary and Halawi, Danny and Zhang, Fred and Tetlock, Philip E},
  booktitle={International Conference on Learning Representations},
  volume={2025},
  pages={93943--93980},
  year={2025}
}

@article{turtel2025,
  title={{{LLMs} can teach themselves to better predict the future}},
  author={Turtel, Benjamin and Franklin, Danny and Schoenegger, Philipp},
  journal={arXiv preprint arXiv:2502.05253},
  year={2025}
}

@article{yang2020,
  title={{{FinBERT}: A pretrained language model for financial communications}},
  author={Yang, Yi and Uy, Mark Christopher Siy and Huang, Allen},
  journal={arXiv preprint arXiv:2006.08097},
  year={2020}
}

@article{xie2024,
  title={{{FinBen}: A holistic financial benchmark for Large Language Models}},
  author={Xie, Qianqian and Han, Weiguang and Chen, Zhengyu and Xiang, Ruoyu and Zhang, Xiao and He, Yueru and Xiao, Mengxi and Li, Dong and Dai, Yongfu and Feng, Duanyu and others},
  journal={Advances in Neural Information Processing Systems},
  volume={37},
  pages={95716--95743},
  year={2024}
}

@inproceedings{mohammad2016,
  title={{{SemEval-2016} task 6: Detecting Stance in Tweets}},
  author={Mohammad, Saif and Kiritchenko, Svetlana and Sobhani, Parinaz and Zhu, Xiaodan and Cherry, Colin},
  booktitle={Proceedings of the 10th International Workshop on Semantic Evaluation (SemEval-2016)},
  pages={31--41},
  year={2016}
}

@inproceedings{rashkin2018,
  title={{{Event2Mind}: Commonsense Inference on Events, Intents, and Reactions}},
  author={Rashkin, Hannah and Sap, Maarten and Allaway, Emily and Smith, Noah A and Choi, Yejin},
  booktitle={Proceedings of the 56th Annual Meeting of the Association for Computational Linguistics (Volume 1: Long Papers)},
  pages={463--473},
  year={2018}
}

@inproceedings{sap2019,
  title={{Social {IQa}: Commonsense Reasoning about Social Interactions}},
  author={Sap, Maarten and Rashkin, Hannah and Chen, Derek and Le Bras, Ronan and Choi, Yejin},
  booktitle={Proceedings of the 2019 conference on Empirical Methods in Natural Language Processing and the 9th International Joint Conference on Natural Language Processing (EMNLP-IJCNLP)},
  pages={4463--4473},
  year={2019}
}

@inproceedings{hutto2014,
  title={{{VADER}: A Parsimonious Rule-Based Model for Sentiment Analysis of Social Media Text}},
  author={Hutto, Clayton and Gilbert, Eric},
  booktitle={Proceedings of the International AAAI Conference on Web and Social Media},
  volume={8(1)},
  pages={216--225},
  year={2014}
}

@article{buz2024,
  title={{{WallStreetBets}: Assessing the Collective Intelligence of {Reddit} for Investment Advice}},
  author={Buz, Tolga and de Melo, Gerard},
  journal={ACM Transactions on Social Computing},
  volume={7},
  number={1-4},
  pages={1--23},
  year={2024},
  publisher={ACM New York, NY}
}

@article{lamorgia2023,
  title={{The Doge of Wall Street: Analysis and Detection of Pump and Dump Cryptocurrency Manipulations}},
  author={La Morgia, Massimo and Mei, Alessandro and Sassi, Francesco and Stefa, Julinda},
  journal={ACM Transactions on Internet Technology},
  volume={23},
  number={1},
  pages={1--28},
  year={2023},
  publisher={ACM New York, NY}
}

@misc{openai2025gpt55,
  author       = {{OpenAI}},
  title        = {{{GPT-5.5} System Card}},
  year         = {2026},
  howpublished = {\url{https://openai.com/index/gpt-5-5-system-card/}},
  note         = {Accessed: 2026-05-04}
}

@misc{anthropic2025haiku45,
  author       = {{Anthropic}},
  title        = {{{Claude Haiku 4.5} System Card}},
  year         = {2025},
  howpublished = {\url{https://www-cdn.anthropic.com/7aad69bf12627d42234e01ee7c36305dc2f6a970.pdf}},
  note         = {Accessed: 2026-05-04}
}

@article{grattafiori2024llama,
  title={{The Llama 3 Herd of Models}},
  author={Grattafiori, Aaron and Dubey, Abhimanyu and Jauhri, Abhinav and Pandey, Abhinav and Kadian, Abhishek and Al-Dahle, Ahmad and Letman, Aiesha and Mathur, Akhil and Schelten, Alan and Vaughan, Alex and others},
  journal={arXiv preprint arXiv:2407.21783},
  year={2024}
}

@misc{mistral7b,
      title={{Mistral 7B}}, 
      author={Albert Q. Jiang and Alexandre Sablayrolles and Arthur Mensch and Chris Bamford and Devendra Singh Chaplot and Diego de las Casas and Florian Bressand and Gianna Lengyel and Guillaume Lample and Lucile Saulnier and Lélio Renard Lavaud and Marie-Anne Lachaux and Pierre Stock and Teven Le Scao and Thibaut Lavril and Thomas Wang and Timothée Lacroix and William El Sayed},
      year={2023},
      eprint={2310.06825},
      archivePrefix={arXiv},
      primaryClass={cs.CL},
      url={https://arxiv.org/abs/2310.06825}, 
}

@inproceedings{
    cheng2024finchat,
    title={{Adapting Large Language Models via Reading Comprehension}},
    author={Cheng, Daixuan and Huang, Shaohan and Wei, Furu},
    booktitle={The Twelfth International Conference on Learning Representations},
    year={2024},
    url={https://openreview.net/forum?id=y886UXPEZ0}
}

@article{xie2023pixiu,
  title={{{PIXIU}: A Comprehensive Benchmark, Instruction Dataset and Large Language Model for Finance}},
  author={Xie, Qianqian and Han, Weiguang and Zhang, Xiao and Lai, Yanzhao and Peng, Min and Lopez-Lira, Alejandro and Huang, Jimin},
  journal={Advances in Neural Information Processing Systems},
  volume={36},
  pages={33469--33484},
  year={2023}
}

@article{comanici2025gemini,
  title={{Gemini 2.5: Pushing the Frontier with Advanced Reasoning, Multimodality, Long Context, and Next Generation Agentic Capabilities}},
  author={Comanici, Gheorghe and Bieber, Eric and Schaekermann, Mike and Pasupat, Ice and Sachdeva, Noveen and Dhillon, Inderjit and Blistein, Marcel and Ram, Ori and Zhang, Dan and Rosen, Evan and others},
  journal={arXiv preprint arXiv:2507.06261},
  year={2025}
}

@article{qian2025fino1,
  title={{Fino1: On the Transferability of Reasoning-Enhanced {LLMs} and Reinforcement Learning to Finance}},
  author={Qian, Lingfei and Zhou, Weipeng and Wang, Yan and Peng, Xueqing and Yi, Han and Zhao, Yilun and Huang, Jimin and Xie, Qianqian and Nie, Jian-yun},
  journal={arXiv preprint arXiv:2502.08127},
  year={2025}
}

@article{yang2025qwen3,
  title={{Qwen3 Technical Report}},
  author={Yang, An and Li, Anfeng and Yang, Baosong and Zhang, Beichen and Hui, Binyuan and Zheng, Bo and Yu, Bowen and Gao, Chang and Huang, Chengen and Lv, Chenxu and others},
  journal={arXiv preprint arXiv:2505.09388},
  year={2025}
}

@article{team2024gemma,
  title={{Gemma 2: Improving Open Language Models at a Practical Size}},
  author={Team, Gemma and Riviere, Morgane and Pathak, Shreya and Sessa, Pier Giuseppe and Hardin, Cassidy and Bhupatiraju, Surya and Hussenot, L{\'e}onard and Mesnard, Thomas and Shahriari, Bobak and Ram{\'e}, Alexandre and others},
  journal={arXiv preprint arXiv:2408.00118},
  year={2024}
}

@article{guo2025deepseek,
  title={{DeepSeek-R1: Incentivizing Reasoning Capability in {LLMs} via Reinforcement Learning}},
  author={Guo, Daya and Yang, Dejian and Zhang, Haowei and Song, Junxiao and Wang, Peiyi and Zhu, Qihao and Xu, Runxin and Zhang, Ruoyu and Ma, Shirong and Bi, Xiao and others},
  journal={arXiv preprint arXiv:2501.12948},
  year={2025}
}

@article{zou2022autocast,
  title={{Forecasting Future World Events with Neural Networks}},
  author={Zou, Andy and Xiao, Tristan and Jia, Ryan and Kwon, Joe and Mazeika, Mantas and Li, Richard and Song, Dawn and Steinhardt, Jacob and Evans, Owain and Hendrycks, Dan},
  journal={Advances in Neural Information Processing Systems},
  volume={35},
  pages={27293--27305},
  year={2022}
}

@inproceedings{zhang2024reality,
  title={{Sentiment Analysis in the Era of Large Language Models: A Reality Check}},
  author={Zhang, Wenxuan and Deng, Yue and Liu, Bing and Pan, Sinno and Bing, Lidong},
  booktitle={Findings of the Association for Computational Linguistics: NAACL 2024},
  pages={3881--3906},
  year={2024}
}

@inproceedings{kheiri2023llms,
  title={{{LLMs} to the Moon? {Reddit} Market Sentiment Analysis with Large Language Models}},
  author={Deng, Xiang and Bashlovkina, Vasilisa and Han, Feng and Baumgartner, Simon and Bendersky, Michael},
  booktitle={Companion Proceedings of the ACM Web Conference 2023},
  pages={1014--1019},
  year={2023}
}

@article{benjamini1995controlling,
  title={Controlling the false discovery rate: a practical and powerful approach to multiple testing},
  author={Benjamini, Yoav and Hochberg, Yosef},
  journal={Journal of the Royal Statistical Society: Series B (Methodological)},
  volume={57},
  number={1},
  pages={289--300},
  year={1995},
  publisher={Wiley Online Library}
}

@incollection{efron1992bootstrap,
  title={{Bootstrap Methods: Another Look at the Jackknife}},
  author={Efron, Bradley},
  booktitle={Breakthroughs in Statistics: Methodology and Distribution},
  pages={569--593},
  year={1992},
  publisher={Springer}
}

@article{malo2014good,
  title={{Good Debt or Bad Debt: Detecting Semantic Orientations in Economic Texts}},
  author={Malo, Pekka and Sinha, Ankur and Korhonen, Pekka and Wallenius, Jyrki and Takala, Pyry},
  journal={Journal of the Association for Information Science and Technology},
  volume={65},
  number={4},
  pages={782--796},
  year={2014},
  publisher={Wiley Online Library}
}

@inproceedings{sclar2024quantifying,
  title={{Quantifying Language Models' Sensitivity to Spurious Features in Prompt Design or: How {I} Learned to Start Worrying about Prompt Formatting}},
  author={Sclar, Melanie and Choi, Yejin and Tsvetkov, Yulia and Suhr, Alane},
  booktitle={International Conference on Learning Representations},
  volume={2024},
  pages={25055--25083},
  year={2024}
}

\appendix


\newpage
\section{Related Work}
\label{sec:related}

\textbf{Forecasting and prediction-market benchmarks.}
Recent forecasting benchmarks evaluate whether models can assign calibrated
probabilities to future events. Autocast~\citep{zou2022autocast} introduced
large-scale event forecasting from textual evidence, \citet{halawi2024}
evaluate LLM calibration on resolved forecasting questions, and
ForecastBench~\citep{karger2025} maintains a dynamic leaderboard for AI
forecasting capabilities. \citet{turtel2025} further use
prediction-market-style forecasting data for model self-improvement.
These works ask whether models predict event outcomes. StakeBench instead
asks whether models understand what a speaker's market comment reveals about
their verified position, later action, and collective odds-direction signal.

\textbf{Financial NLP, sentiment, and stance.}
Financial NLP has largely focused on document- or sentence-level labels.
FinBERT~\citep{yang2020} adapts language modeling to financial communications,
PIXIU~\citep{xie2023pixiu} combines a benchmark, instruction dataset, and
financial LLM, and FinBen~\citep{xie2024} broadens evaluation across financial
tasks. General sentiment and stance resources, including VADER~\citep{hutto2014},
SemEval-2016 Task~6~\citep{mohammad2016}, and the multi-dataset LLM sentiment
audit of \citet{zhang2024reality}, evaluate how text is classified by external
labels. SemEval's YES/NO/NONE formulation is structurally close to StakeBench
\(\mathsf{G}_2\), but its labels are crowdsourced tweet stances rather than positions that
speakers paid to hold. StakeBench replaces perception-based labels with
revealed-preference targets~\citep{samuelson1948}.

\textbf{Market discourse and social-media finance.}
Work on retail-investor communities studies how online language relates to
market behavior. \citet{kheiri2023llms} analyze Reddit market sentiment with
LLMs, \citet{buz2024} study WallStreetBets investment advice, and
\citet{lamorgia2023} analyze pump-and-dump manipulation in cryptocurrency
communities. These studies connect public discussion to market phenomena, but
they typically operate at the post, forum, or aggregate level. StakeBench links
each comment to the author's own contemporaneous position and post-comment
trading record, enabling instance-level tests of commitment language rather
than only community-level sentiment analysis.

\textbf{Grounded intent, action, and strategic communication.}
NLP benchmarks for social reasoning, such as Event2Mind~\citep{rashkin2018} and
Social IQa~\citep{sap2019}, test whether models can infer intents, reactions, and
likely social consequences from text, but their targets are human-authored
annotations. StakeBench \(\mathsf{G}_3\) extends grounded intent prediction to observed
post-comment market actions. The benchmark is also connected to economic
accounts of costly signaling and strategic communication: Spence's signaling
model~\citep{spence1978job} explains why costly actions can reveal private
information, while Crawford and Sobel's cheap-talk framework~\citep{crawford1982}
shows why unsupported messages can be strategically distorted. Prediction-market
positions provide the missing costly action, allowing StakeBench to test
whether models distinguish committed speech from unsupported market talk.


\section{Detailed settings}
\label{app:detailed settings}
\textbf{Models.}
The model suite is used as a set of probes for the benchmark rather than as the
main contribution. The goal is to expose capability boundaries under a fixed
prompting protocol, not to exhaust model-specific prompt tuning or establish a
final model ranking. All 15 models use greedy decoding
(temperature\,${=}$\,0, seed\,${=}$\,42).
Three closed-source models are queried via the OpenRouter API:
GPT-5.5~\citep{openai2025gpt55}, Gemini-2.5-Flash~\citep{comanici2025gemini},
and Claude-Haiku-4.5~\citep{anthropic2025haiku45}.
Eight general open-weight models are loaded from Hugging Face in float16:
Qwen3-32B, Qwen3-14B, Qwen3-30B-A3B (mixture-of-experts, MoE), and
Qwen3-8B~\citep{yang2025qwen3},
Gemma2-9B~\citep{team2024gemma},
DeepSeek-R1-Distill-Llama-8B~\citep{guo2025deepseek},
Llama-3.2-3B-Instruct~\citep{grattafiori2024llama}, and
Mistral-7B-Instruct-v0.3~\citep{mistral7b}.
Four finance-domain models complete the suite:
Fino1-14B (Qwen2.5-14B base) and Fino1-8B (Llama-3.1-8B
base)~\citep{qian2025fino1},
Finance-Chat-7B (Llama-2-7B-Chat base)~\citep{cheng2024finchat}, and
FinMA-30B (Llama-2-30B base)~\citep{xie2023pixiu}.
Qwen3 models prepend the \texttt{<nothink>} token to suppress chain-of-thought
generation and are capped at 32 output tokens.
DeepSeek-R1-8B retains its full reasoning chain (512 output tokens).
All other open-weight models are capped at 24--48 output tokens.
All 15 models are evaluated on the same 18 topic--platform splits.

\textbf{Computational environment.}
Open-weight models are evaluated on a SLURM-managed high-performance-computing
(HPC) cluster equipped with NVIDIA GH200 GPUs (95\,GB HBM3e per device), using
the Hugging Face Transformers library (Python~3.11.7, PyTorch~2.5.1, float16
precision).
Most models fit on a single GH200. Qwen3-30B-A3B (MoE, ${\approx}$60\,GB fp16)
and FinMA-30B (${\approx}$60\,GB fp16) use tensor parallelism
(\texttt{device\_map="auto"}) across two GPUs.
Per-model inference batch sizes are calibrated to the available key--value (KV)
cache budget on one GH200 at a 1{,}500-token context. Batch sizes range from
32 prompts for Qwen3-32B (64\,GB weights, ${\approx}$12\,GB KV headroom) to 384
for Llama-3.2-3B (6.4\,GB weights, ${\approx}$63\,GB KV headroom).
Closed-source responses are logged and parsed by the same label-extraction
pipeline used for open-weight models.

\textbf{Task windows and filters.}
\(\mathsf{G}_3\) uses the post-comment action window defined in
Section~\ref{sec:notation}. \(\mathsf{G}_4\) uses an adaptive per-window horizon
\(h\in[1,30]\) days. Topic--platform splits
with fewer than 12 instances are excluded, and \(\mathsf{G}_4\) excludes near-zero odds moves
with \(|p_w^h-p_w^0|\leq0.005\).

\textbf{Scoring and significance.}
For platform comparisons, \(\delta\) denotes the paired Polymarket-minus-Manifold
mean score difference over shared topics, and \(\rho\) denotes the corresponding
Spearman rank correlation of model rankings across platforms. Cross-platform
breakdowns are reported in Appendix Table~\ref{tab:aggregate-analysis}. Macro averages weight all 18 topic--platform splits equally. Significance uses one-sided binomial tests (\(\alpha=0.05\)) with null
baselines 0.5 for \(\mathsf{G}_1\), \(\mathsf{G}_2\)-DA, and the binary
\(\mathsf{G}_3\) slices \(\mathrm{Acc}_f\) and \(\mathrm{Acc}_d\), and 0.25 for
\(\mathsf{G}_3\)-\(\mathrm{Acc}_{\mathrm{act}}\).
\(\mathsf{G}_4\) PLA and \(n_+\) are reported descriptively without a
significance threshold. The aggregate score
\(\mathrm{Agg}\) rescales \(\mathsf{G}_1\)-Acc, \(\mathsf{G}_2\)-DA, and
\(\mathsf{G}_3\)-\(\mathrm{Acc}_{\mathrm{act}}\) from their chance baselines
\((0.5,0.5,0.25)\) toward 1, rescales \(n_+\) from 0 toward 18, clips each
component below at 0, and averages the four components equally. Plain \(p\)
denotes a statistical test p-value. Spearman \(\rho\) is a signed
rank correlation, while \(|\rho|\) reports only its magnitude when the claim
concerns the absence of association regardless of direction. Bootstrap
confidence interval widths~\citep{efron1992bootstrap}
and Benjamini-Hochberg false discovery rate retention~\citep{benjamini1995controlling}
are reported in Appendix~\ref{app:audits}.

\textbf{Prompting.}
\(\mathsf{G}_1\) uses a one-shot prompt requesting YES or NO.
\(\mathsf{G}_2\) requests one label from the prediction set in
Table~\ref{tab:tasks} and
instructs the model to infer the commenter's revealed position side rather than
the market outcome or surface sentiment. \(\mathsf{G}_3\) provides \(z_i\) and
\(s_i\), then requests the four-class action label or a binary subtask label.
\(\mathsf{G}_4\) provides \(p_w^0\), concatenates window
comments, and requests one direction label from Table~\ref{tab:tasks}.
Full prompt templates are in Appendix~\ref{app:prompts}.

\section{Statistical Reliability and Validity Audit Details}
\label{app:audits}

\begin{table}[tbh]
\centering
\caption{Statistical reliability checks. \(N^{\mathrm{pool}}_+\) is the pooled
\(\mathsf{G}_4\) positive count, distinct from the per-model \(n_+\) in
Table~\ref{tab:main}; Ret.\ is the fraction of raw discoveries retained after
BH-FDR correction.}
\label{tab:eval-checks}
\footnotesize
\setlength{\tabcolsep}{5pt}
\renewcommand{\arraystretch}{1.08}
\begin{tabular*}{\linewidth}{@{\extracolsep{\fill}}lrrrr@{}}
\toprule
\rowcolor{tablehead}
\textbf{Metric} & \textbf{95\% CI} & \textbf{Raw sig.} & \textbf{BH-FDR} & \textbf{Ret.} \\
\midrule
\(\mathsf{G}_1\) Acc      & .033 & 166 & 154 & 92.8\% \\
\(\mathsf{G}_2\) DA       & .050 & 114 &  72 & 63.2\% \\
\(\mathsf{G}_3\) \(\mathrm{Acc}_{\mathrm{act}}\)  & .057 &  17 &   7 & 41.2\% \\
\(\mathsf{G}_3\) \(\mathrm{Acc}_f\)  & .096 &  23 &   7 & 30.4\% \\
\addlinespace[1pt]
\rowcolor{tablesubhead}
\(\mathsf{G}_4\) PLA & .116 & \multicolumn{3}{c@{}}{\textit{$N^{\mathrm{pool}}_+\,=\,32\,/\,270\enskip(\Delta_{\mathrm{PLA}}{>}0)$}} \\
\bottomrule
\end{tabular*}
\end{table}
Reliability declines as tasks move from market commitment detection to
behavioral anticipation. \(\mathsf{G}_1\) and \(\mathsf{G}_2\) retain most discoveries after
BH-FDR correction, while \(\mathsf{G}_3\) has wider uncertainty and fewer robust
topic-level wins.
The \(n_+\) column in Table~\ref{tab:main} is per-model rather than pooled.

\begin{table}[tbh]
\centering
\caption{Action-window sensitivity. Sig.\ splits are model--split tests above
the chance baseline.}
\label{tab:g3-window-sensitivity}
\footnotesize
\setlength{\tabcolsep}{5pt}
\renewcommand{\arraystretch}{1.08}
\begin{tabular*}{\linewidth}{@{\extracolsep{\fill}}lrrrr@{}}
\toprule
\rowcolor{tablehead}
\textbf{Horizon} & \textbf{True hold} & \textbf{Pred. hold} &
\(\boldsymbol{\mathrm{Mean\ Acc}_{\mathrm{act}}}\) & \textbf{Sig. splits} \\
\midrule
6h  & 45.3\% & 22.1\% & .233 & 21 \\
12h & 45.3\% & 22.1\% & .231 & 18 \\
24h & 45.2\% & 22.6\% & .229 & 18 \\
48h & 45.1\% & 22.5\% & .231 & 21 \\
72h & 45.1\% & 22.4\% & .228 & 15 \\
\bottomrule
\end{tabular*}
\end{table}
Mean accuracy varies by less than one percentage point across horizons, and the
ground-truth hold rate is nearly unchanged. The final column counts
model--split tests above the chance baseline.

\textbf{Full audit findings.}
In the Wrong-Bettor test, 25 of 116 eligible model--split pairs reach $p{<}0.05$, five of which survive Benjamini--Hochberg correction at \(q<0.05\) (Appendix~\ref{app:bh}). On \textsc{poly/politics\_intl}, $\Pr(\hat{z}_i{=}z_i\mid
z_i{\neq}r_i)$ reaches $0.71$--$0.81$ across four models, showing that
models track commitment language rather than market truth.
In the stake-magnitude dose-response audit, 12 of 154 eligible pairs are
significant, with five concentrated on \textsc{mani/economics}
(CSD\({\approx}0.26\)--\(0.37\)). Evidence outside this topic is
limited.
The formal DiD statistic ($\mathrm{DA}^+_r - \mathrm{DA}^-_r$, both scored
against $r_i$) has macro mean $0.026$.
As a complementary comparison, $\mathrm{DA}^+_z$ (post-position scored against
\(z_i\))
reaches $0.621$ versus $\mathrm{DA}^-_r{=}0.508$,
indicating stronger commitment-side signal after position-taking.

Tables~\ref{tab:wrong-bettor}, \ref{tab:dose-response}, and
\ref{tab:did-permodel} give the full per-model breakdowns for the three audits
on all 12 open-weight models
(116 eligible and 25 significant pairs for the Wrong-Bettor test; 3 significant dose-response cases).

\begin{table}[tbh]
\centering
\caption{Wrong-Bettor test summary. Elig.\ counts eligible topic--model pairs,
Sig.\ counts significant tests, and Mac.\ pct is the macro commitment-following
rate.}
\label{tab:wrong-bettor}
\footnotesize
\setlength{\tabcolsep}{4pt}
\renewcommand{\arraystretch}{1.06}
\begin{tabular*}{\linewidth}{@{\extracolsep{\fill}}lrrrr@{}}
\toprule
\rowcolor{tablehead}
\textbf{Model} & \textbf{Elig.} & \textbf{Sig.} & $\boldsymbol{>0.5}$ & \textbf{Mac.\ pct} \\
\midrule
Qwen3-8B        &  8 & 4 & 5 & .584 \\
Qwen3-32B       & 10 & 3 & 6 & .552 \\
Qwen3-14B       & 10 & 3 & 5 & .535 \\
DeepSeek-R1-8B  &  9 & 1 & 6 & .533 \\
Fin-Chat-7B     & 10 & 1 & 5 & .526 \\
Fino1-14B       &  8 & 2 & 4 & .505 \\
Gemma2-9B       &  9 & 3 & 5 & .504 \\
Mistral-7B      & 11 & 1 & 5 & .474 \\
Qwen3-30B       &  8 & 2 & 5 & .472 \\
Fino1-8B        & 11 & 3 & 5 & .462 \\
Llama-3.2-3B    & 11 & 0 & 4 & .432 \\
FinMA-30B       & 11 & 2 & 4 & .384 \\
\addlinespace[1pt]
\rowcolor{tablesubhead}
\textit{Total}  & 116 & 25 & 59 & .493 \\
\bottomrule
\end{tabular*}
\end{table}

In Table~\ref{tab:wrong-bettor}, \(>0.5\) counts eligible model--topic pairs
whose commitment-following rate is above chance. The strongest effects come
from a small set of topics rather than a uniform global shift.

\begin{table}[tbh]
\centering
\caption{Outcome-scored side audit. Gap is resolution-scored DA minus
side-scored DA.}
\label{tab:outcome-side-audit}
\footnotesize
\setlength{\tabcolsep}{4pt}
\renewcommand{\arraystretch}{1.06}
\begin{tabular*}{\linewidth}{@{\extracolsep{\fill}}llrrrr@{}}
\toprule
\rowcolor{tablehead}
\textbf{Type} & \textbf{Model} & \textbf{Topics} &
\textbf{Side DA} & \textbf{Res. DA} & \textbf{Gap} \\
\midrule
\rowcolor{tablesubhead}
general & DeepSeek-R1-8B & 18 & .514 & .497 & $-$.018 \\
general & Gemma2-9B      & 18 & .578 & .567 & $-$.011 \\
general & Llama-3.2-3B   & 18 & .534 & .550 & $+$.015 \\
general & Mistral-7B     & 18 & .523 & .521 & $-$.003 \\
general & Qwen3-14B      & 18 & .567 & .531 & $-$.036 \\
general & Qwen3-30B      & 18 & .548 & .543 & $-$.004 \\
general & Qwen3-32B      & 18 & .578 & .547 & $-$.031 \\
general & Qwen3-8B       & 18 & .564 & .530 & $-$.035 \\
\addlinespace[1pt]
\rowcolor{tablesubhead}
finance & Fin-Chat-7B & 18 & .514 & .510 & $-$.004 \\
finance & Fino1-14B       & 18 & .565 & .556 & $-$.009 \\
finance & Fino1-8B        & 18 & .524 & .522 & $-$.002 \\
finance & FinMA-30B       & 18 & .506 & .538 & $+$.032 \\
\bottomrule
\end{tabular*}
\end{table}
FinMA-30B is the only model with a clear positive gap, suggesting that its side
predictions lean toward
market outcomes rather than commenter-side language.

\begin{table}[tbh]
\centering
\caption{Stake-magnitude dose response.}
\label{tab:dose-response}
\footnotesize
\setlength{\tabcolsep}{3.5pt}
\renewcommand{\arraystretch}{1.06}
\begin{tabular*}{\linewidth}{@{\extracolsep{\fill}}llrrrrrrr@{}}
\toprule
\rowcolor{tablehead}
\textbf{Model} & \textbf{Topic} & $n_d$ & \textbf{CSD} & $p$ &
\(\boldsymbol{\mathrm{DA}^{(1)}}\) & \(\boldsymbol{\mathrm{DA}^{(2)}}\) &
\(\boldsymbol{\mathrm{DA}^{(3)}}\) & \(\boldsymbol{\mathrm{DA}^{(4)}}\) \\
\midrule
FinMA-30B  & mani/economics & 116 & .321 & .017 & .300 & .429 & .759 & .621 \\
Qwen3-30B  & mani/economics &  75 & .368 & .026 & .158 & .579 & .722 & .526 \\
Qwen3-8B   & mani/economics &  70 & .333 & .044 & .333 & .706 & .765 & .667 \\
\bottomrule
\end{tabular*}
\end{table}
All significant dose-response cases occur on Manifold economics. The repeated
pattern suggests that stake size can sharpen commitment language, but the
evidence is topic-concentrated rather than universal.
Here $n_d$ is the number of directional examples used in the quartile test
(consistent with the notation in Section~\ref{sec:eval-protocol}).

\begin{table}[tbh]
\centering
\caption{Within-user DiD summary. Neg./total counts topics with negative DiD.}
\label{tab:did-permodel}
\footnotesize
\setlength{\tabcolsep}{3.5pt}
\renewcommand{\arraystretch}{1.06}
\begin{tabular*}{\linewidth}{@{\extracolsep{\fill}}lrrrr@{}}
\toprule
\rowcolor{tablehead}
\textbf{Model} & \textbf{Macro DiD} & \textbf{Neg./total} &
\makecell{\textbf{Pre}\\\textbf{vs. }$r_i$} &
\makecell{\textbf{Post}\\\textbf{vs. }$z_i$} \\
\midrule
Fin-Chat-7B     & $+$0.075 & 3/12 & .453 & .548 \\
FinMA-30B       & $+$0.066 & 4/12 & .495 & .547 \\
Qwen3-30B       & $+$0.052 & 3/12 & .470 & .639 \\
Llama-3.2-3B    & $+$0.046 & 5/12 & .497 & .560 \\
Gemma2-9B       & $+$0.034 & 4/12 & .523 & .689 \\
Qwen3-8B        & $+$0.020 & 5/12 & .509 & .677 \\
Mistral-7B      & $+$0.016 & 6/12 & .541 & .675 \\
DeepSeek-R1-8B  & $+$0.008 & 6/12 & .495 & .543 \\
Qwen3-32B       & $-$0.030 & 6/12 & .556 & .656 \\
Qwen3-14B       & $-$0.029 & 7/12 & .542 & .678 \\
\addlinespace[1pt]
\rowcolor{tablesubhead}
\textit{Mean}   & $+$0.026 & 4.9/12 & .508 & .621 \\
\bottomrule
\end{tabular*}
\end{table}
Macro DiD is the macro-averaged formal resolution-scored statistic
($\mathrm{DA}^+_r-\mathrm{DA}^-_r$).
The final two columns show the complementary comparison $\mathrm{DA}^-_r$ and
$\mathrm{DA}^+_z$, confirming that
post-position language aligns more strongly with the user's own side than
pre-position language aligns with the final market outcome.

\section{Robustness Analyses}
\label{app:robust}

\subsection{Lexical Recoverability Probe}
\label{app:probe}

A TF-IDF bag-of-words representation with logistic regression is trained on
comment text only, without market, position or user features, on the union
of released evaluation instances, separately per platform and tier. The
chronological split places the earliest 70\% of comments in training, so
every training comment precedes every test comment. The market-disjoint and
commenter-disjoint splits assign whole markets or whole commenters to one
side in a single random 70/30 assignment. The statistic is balanced accuracy
against the \(1/k\) class floor, which a constant predictor attains exactly,
with a 1{,}000-draw label-permutation test. Instances whose gold label
differs between model logs are excluded: none for \(\mathsf{G}_1\) and
\(\mathsf{G}_2\), and 89 of 3{,}562 (Manifold) and 18 of 1{,}624
(Polymarket) for \(\mathsf{G}_3\).

\begin{table}[tbh]
\centering
\caption{Lexical recoverability probe: balanced accuracy of a TF-IDF
logistic-regression probe trained on comment text only, under a chronological
split and two leakage-controlled splits, with label-permutation
\(p\)-values (1{,}000 draws). Floor is \(1/k\).}
\label{tab:probe}
\footnotesize
\setlength{\tabcolsep}{4pt}
\renewcommand{\arraystretch}{1.06}
\begin{tabular*}{\linewidth}{@{\extracolsep{\fill}}llrrrrrrrr@{}}
\toprule
\rowcolor{tablehead}
& & & & \multicolumn{2}{c}{\textbf{Chronological}} & \multicolumn{2}{c}{\textbf{Market-disjoint}} & \multicolumn{2}{c}{\textbf{Commenter-disjoint}} \\
\cmidrule(lr){5-6}\cmidrule(lr){7-8}\cmidrule(l){9-10}
\rowcolor{tablesubhead}
\textbf{Platform} & \textbf{Tier} & \(\boldsymbol{n}\) & \textbf{Floor} & \textbf{Bal-Acc} & \(\boldsymbol{p}\) & \textbf{Bal-Acc} & \(\boldsymbol{p}\) & \textbf{Bal-Acc} & \(\boldsymbol{p}\) \\
\midrule
Manifold   & \(\mathsf{G}_1\) &   616 & .500 & .685 & .001 & .657 & .001 & .667 & .001 \\
Manifold   & \(\mathsf{G}_2\) & 1,569 & .500 & .531 & .145 & .526 & .153 & .544 & .035 \\
Manifold   & \(\mathsf{G}_3\) & 3,470 & .250 & .246 & .639 & .257 & .300 & .282 & .007 \\
\addlinespace[1pt]
Polymarket & \(\mathsf{G}_1\) &   799 & .500 & .729 & .001 & .683 & .001 & .701 & .001 \\
Polymarket & \(\mathsf{G}_2\) & 1,740 & .500 & .521 & .205 & .462 & .962 & .550 & .015 \\
Polymarket & \(\mathsf{G}_3\) & 1,606 & .250 & .228 & .838 & .248 & .540 & .232 & .814 \\
\bottomrule
\end{tabular*}
\end{table}

\textbf{Reading.} \(\mathsf{G}_3\) recoverability is small and
split-dependent rather than exactly zero: the Manifold commenter-disjoint
split reaches 0.282 (\(p{=}0.007\)), while the other five \(\mathsf{G}_3\)
cells stay within 0.03 of the floor. A linear bag-of-words probe cannot rule
out non-linear, discourse-level or market-conditioned signal, so a low score
means low shallow lexical recoverability and nothing stronger. Balanced
accuracy here counts NEUTRAL as an error and is not the paper's Directed
Accuracy.

\subsection{Wrong-Bettor Audit with Multiplicity Correction}
\label{app:bh}

Across the 116 eligible model--topic tests of the Wrong-Bettor audit
(Table~\ref{tab:wrong-bettor}), 25 reach uncorrected \(p{<}0.05\). Five
survive Benjamini--Hochberg correction at \(q{<}0.05\)
(Table~\ref{tab:bh-cells}), and the same five survive a Bonferroni threshold
of \(0.05/116\). The macro mean commitment-following rate is 0.493 (median
0.513, with 59 of 116 cells above 0.5), so the aggregate is at chance and no
global claim is made. Four of the five surviving cells fall in Polymarket
politics\_intl, so commitment-following is localized where it appears. The
peak cell (0.844) should be read together with this correction rather than
alone.

\begin{table}[tbh]
\centering
\caption{Wrong-Bettor cells surviving Benjamini--Hochberg correction
(\(q{<}0.05\)). Follow-rate is the share of directional predictions matching
the commenter's (losing) side.}
\label{tab:bh-cells}
\footnotesize
\setlength{\tabcolsep}{5pt}
\renewcommand{\arraystretch}{1.06}
\begin{tabular*}{\linewidth}{@{\extracolsep{\fill}}llrrrr@{}}
\toprule
\rowcolor{tablehead}
\textbf{Model} & \textbf{Platform / topic} & \textbf{Follow-rate} & \(\boldsymbol{n}\) & \textbf{raw} \(\boldsymbol{p}\) & \textbf{BH} \(\boldsymbol{q}\) \\
\midrule
Qwen3-8B  & Manifold / other            & .844 & 32 & .0001 & .0029 \\
Qwen3-30B & Polymarket / politics\_intl & .806 & 67 & $<$.0001 & $<$.0001 \\
FinMA-30B & Polymarket / politics\_intl & .743 & 74 & $<$.0001 & $<$.0001 \\
Fino1-14B & Polymarket / politics\_intl & .721 & 61 & .0004 & .0093 \\
Fino1-8B  & Polymarket / politics\_intl & .709 & 79 & .0001 & .0029 \\
\bottomrule
\end{tabular*}
\end{table}

\subsection{Activity-Skew Stress Test on Manifold}
\label{app:skew}

Manifold's position coverage (80.5\%) makes the full trading distribution of
its commenters observable, so an activity restriction can be imposed there
and its effect on the \(\mathsf{G}_3\) result measured. Two restrictions
answer different questions. Restriction (B) ranks the 853 unique Manifold
commenters that appear in any model's \(\mathsf{G}_3\) log by lifetime bet
count and keeps the top 14.2\% (121 commenters), the same set for every
model. Because active commenters comment more, this keeps 18.9\% of an
average model's commenters and 39.8\% of its \(\mathsf{G}_3\) instances.
Restriction (A) keeps the top \(\approx\)14\% of comment instances per model
by the commenter's bet count, which retains only \(\approx\)2.9\% of
commenters. Every number is the mean over the 15 models of that model's own
quantity, with differences computed per model and then averaged. Confidence
intervals come from a hierarchical commenter-clustered bootstrap (2{,}000
draws) that resamples commenters with replacement and recomputes both
restrictions inside each draw.

\begin{table}[tbh]
\centering
\caption{\(\mathsf{G}_3\) on Manifold under activity restrictions (mean over
15 models). Collapse is the share of predictions on the most frequent label.
Differences are relative to the full set, with commenter-clustered bootstrap
95\% CIs.}
\label{tab:skew}
\footnotesize
\setlength{\tabcolsep}{4pt}
\renewcommand{\arraystretch}{1.06}
\begin{tabular*}{\linewidth}{@{\extracolsep{\fill}}lrrrrr@{}}
\toprule
\rowcolor{tablehead}
\textbf{Evaluation set} & \textbf{Acc} & \textbf{Macro-recall} & \textbf{Collapse} & \(\boldsymbol{\Delta}\)\textbf{ macro-recall [CI]} & \(\boldsymbol{\Delta}\)\textbf{ collapse [CI]} \\
\midrule
Full Manifold                                   & .255 & .254 & .655 & -- & -- \\
(B) top 14.2\% of commenters by bets            & .251 & .249 & .673 & $-$.005 [$-$.014, $+$.004] & $+$.018 [$+$.008, $+$.033] \\
(A) top \(\approx\)14\% of comment instances    & .246 & .235 & .670 & $-$.019 [$-$.042, $+$.005] & $+$.015 [$-$.003, $+$.040] \\
\bottomrule
\end{tabular*}
\end{table}

\begin{table}[tbh]
\centering
\caption{\(\mathsf{G}_3\) on Manifold by commenter-activity quartile. Quartiles
are cut once on the 853 commenters by lifetime bet count and mapped back to
their comments (mean over 15 models).}
\label{tab:skew-quartiles}
\footnotesize
\setlength{\tabcolsep}{5pt}
\renewcommand{\arraystretch}{1.06}
\begin{tabular*}{\linewidth}{@{\extracolsep{\fill}}lrrrrrr@{}}
\toprule
\rowcolor{tablehead}
\textbf{Quartile} & \textbf{Commenters} & \textbf{Mean comments} & \textbf{Median bets} & \textbf{Acc} & \textbf{Macro-recall} & \textbf{Collapse} \\
\midrule
Q1 (least active) & 214 & 162 &  19 & .214 & .248 & .631 \\
Q2                & 213 & 233 &  62 & .254 & .263 & .649 \\
Q3                & 213 & 311 & 155 & .275 & .269 & .662 \\
Q4 (most active)  & 213 & 830 & 829 & .257 & .251 & .667 \\
\bottomrule
\end{tabular*}
\end{table}

\textbf{Reading.} Under restriction (B) no model moves macro-recall by more
than 0.03, and macro-recall shows no trend across commenter quartiles (0.248
to 0.269). This is a Manifold stress test under one activity skew, not proof
that all forms of Polymarket selection bias are eliminated: the labels of
Polymarket's unreconstructed records are unobservable, so their
representativeness remains open (Section~\ref{sec:g3}). Statements about
low-activity participants refer to the commenter quartiles, not to the
instance quantile of restriction (A).

\subsection{Window-Level Text Probe for \(\mathsf{G}_4\)}
\label{app:g4probe}

To ask whether window text carries odds-direction signal beyond the revealed
position distribution, we reconstruct the full pool of evaluation windows per
platform (1{,}163 Manifold, 557 Polymarket) from the released benchmark
files, concatenate the full window text, and train the same TF-IDF
logistic-regression probe under a chronological split. Logistic regression
was fixed as the primary probe in advance, and a gradient-boosted secondary
probe is reported for completeness. Baselines are the majority direction and
the stake-weighted direction \(\tilde d_w\). Paired probe-minus-baseline
differences are computed on windows where both predictors are defined and
bootstrapped over markets (2{,}000 draws, 74 and 41 test markets).

\begin{table}[tbh]
\centering
\caption{Window-level text probe for \(\mathsf{G}_4\) (direction accuracy on
the held-out window set) and paired differences of the logistic-regression
probe against each baseline, market-clustered bootstrap 95\% CIs.}
\label{tab:g4-probe}
\footnotesize
\setlength{\tabcolsep}{4pt}
\renewcommand{\arraystretch}{1.06}
\begin{tabular*}{\linewidth}{@{\extracolsep{\fill}}lrrrrrr@{}}
\toprule
\rowcolor{tablehead}
\textbf{Platform} & \textbf{Windows} & \textbf{Test UP-rate} & \textbf{Majority} & \textbf{Stake-weighted} & \textbf{Probe (LR)} & \textbf{Probe (GBDT)} \\
\midrule
Manifold   & 1,163 & .490 & .510 & .547 & .513 & .536 \\
Polymarket &   557 & .369 & .631 & .530 & .643 & .643 \\
\midrule
\rowcolor{tablesubhead}
\textbf{Platform} & \textbf{Test markets} & \multicolumn{2}{l}{\textbf{LR $-$ majority [95\% CI]}} & \multicolumn{3}{l}{\textbf{LR $-$ stake-weighted [95\% CI]}} \\
\midrule
Manifold   & 74 & \multicolumn{2}{l}{$+$.003 [$-$.083, $+$.089]} & \multicolumn{3}{l}{$-$.034 [$-$.142, $+$.071]} \\
Polymarket & 41 & \multicolumn{2}{l}{$+$.012 [$-$.066, $+$.084]} & \multicolumn{3}{l}{$+$.113 [$+$.006, $+$.229]} \\
\bottomrule
\end{tabular*}
\end{table}

\textbf{Reading.} Against the stronger baseline on each platform
(stake-weighted on Manifold, majority on Polymarket) both paired intervals
contain zero. The only positive interval is the Polymarket probe against the
weaker stake-weighted baseline, and all four comparisons are listed so that
the choice of baseline is visible. The nonlinear secondary probe gives the
same qualitative result. Three boundaries apply. The probe windows are a
separate reconstruction, not the inputs the LLMs saw: the \(\mathsf{G}_4\)
evaluation sampled windows, truncated each to its first ten comments at 120
characters and applied the near-tie and commitment-consensus filters, with
the exact inputs preserved in the released audit logs, whereas this probe
uses the full window pool and full text. No efficiency quantity (spread,
liquidity, absorption speed, calibration) is measured, so the analysis
cannot separate market efficiency from the limits of the inputs the models
received. The platform split of positive \(\mathsf{G}_4\) gains
(Table~\ref{tab:g4-platform-gains}) is not evidence of an efficiency
difference between platforms, since platform, topic composition, window
availability and sample size are confounded. The window pool is regenerated
from the released data by the released script.

\section{Platform Comparison Details}
\label{app:platform}

Here, \(\delta\) denotes the Polymarket-minus-Manifold mean score
difference and \(\rho\) denotes cross-platform Spearman rank correlation.

\begin{table}[tbh]
\centering
\caption{Cross-platform comparison.}
\label{tab:aggregate-analysis}
\scriptsize
\setlength{\tabcolsep}{2pt}
\renewcommand{\arraystretch}{1.05}
\resizebox{0.5\linewidth}{!}{%
\begin{tabular}{lrrrr}
\toprule
\rowcolor{tablehead}
\textbf{Task} & $\boldsymbol{\delta}$ & \textbf{Test stat.} &
$\boldsymbol{p}$ & \textbf{Cross-rank}\,$\boldsymbol{\rho}$ \\
\midrule
\(\mathsf{G}_1\)       & $+$0.016 & \phantom{$-$}0.37 & .729     & $.975^*$ \\
\(\mathsf{G}_2\)-DA    & $+$0.015 & \phantom{$-$}0.44 & .680     & $.286$   \\
\(\mathsf{G}_3\)-\(\mathrm{Acc}_{\mathrm{act}}\) & $-$0.076 & $-$7.06           & $.001^*$ & $.207$   \\
\(\mathsf{G}_3\)-\(\mathrm{Acc}_f\) & $-$0.129 & $-$7.02           & $.001^*$ & $.032$   \\
\(\mathsf{G}_4\) ($\Delta_{\mathrm{PLA}}$) & $-$0.110 & $-$1.04 & .346 & $.771^*$ \\
\bottomrule
\end{tabular}
}
\end{table}
Table~\ref{tab:aggregate-analysis} reports \(\delta\), the paired test statistic,
the test \(p\)-value, and \(\rho\). The main platform effect is concentrated in
\(\mathsf{G}_3\), while \(\mathsf{G}_1\) rankings transfer
almost perfectly.

\begin{table}[tbh]
\centering
\caption{Paired finance-domain comparisons.}
\label{tab:finance-pairs}
\scriptsize
\setlength{\tabcolsep}{2.5pt}
\renewcommand{\arraystretch}{1.06}
\resizebox{\linewidth}{!}{%
\begin{tabular}{llrrrrr}
\toprule
\rowcolor{tablehead}
\textbf{Pair} & \textbf{Comparison} & \(\boldsymbol{\mathsf{G}_1}\) &
\(\boldsymbol{\mathsf{G}_2}\)-\textbf{DA} &
\(\boldsymbol{\mathsf{G}_3}\)-\(\boldsymbol{\mathrm{Acc}_{\mathrm{act}}}\) &
\(\boldsymbol{\mathsf{G}_3}\)-\(\boldsymbol{\mathrm{Acc}_f}\) &
\(\boldsymbol{\mathsf{G}_4}\)-\textbf{PLA} \\
\midrule
14B & Fino1-14B minus Qwen3-14B & $-$.092 & $-$.002 & $-$.084 & $-$.097 & $-$.070 \\
30B & FinMA-30B minus Qwen3-32B & $-$.245 & $-$.072 & $-$.092 & $-$.139 & $-$.272 \\
7B  & Fin-Chat-7B minus Mistral-7B & $-$.004 & $-$.009 & $-$.002 & $+$.090 & $+$.167 \\
\bottomrule
\end{tabular}
}
\end{table}
Entries are finance-model scores minus matched general-model scores.
The paired comparisons show that finance tuning does not improve speaker-level
commitment tasks, even when model sizes are similar.

\begin{table}[tbh]
\centering
\caption{\(\mathsf{G}_4\) platform split.}
\label{tab:g4-platform-gains}
\footnotesize
\setlength{\tabcolsep}{5pt}
\renewcommand{\arraystretch}{1.08}
\begin{tabular*}{\linewidth}{@{\extracolsep{\fill}}lrrrrr@{}}
\toprule
\rowcolor{tablehead}
\textbf{Platform} & \textbf{Splits} & \textbf{Positive} &
\textbf{Share} & \textbf{PLA} & \(\boldsymbol{\Delta}_{\mathrm{PLA}}\) \\
\midrule
Manifold   & 180 & 27 & 15.0\% & .312 & $-$.220 \\
Polymarket &  90 &  5 &  5.6\% & .352 & $-$.300 \\
\addlinespace[1pt]
\rowcolor{tablesubhead}
\textit{Total} & 270 & 32 & 11.9\% & .325 & $-$.247 \\
\bottomrule
\end{tabular*}
\end{table}
Positive counts model--split instances where the model exceeds the naive
odds-direction baseline.
Positive gains are rare on both platforms, but they are more frequent on
Manifold. This supports treating \(\mathsf{G}_4\) as a baseline-relative test of
residual signal rather than a claim that comments alone forecast market odds.

\section{Dataset Details}
\label{app:dataset}

\subsection{Data Sources and Collection Funnel}

The collection design separates breadth from label density. Polymarket provides
large-scale real-money discussion, while Manifold provides cleaner binary
resolution fields and much denser position coverage. We therefore use both
platforms rather than treating either as an interchangeable source.

\begin{table}[tbh]
\centering
\caption{Data sources and collection funnel.}
\label{tab:data-sources-funnel}
\footnotesize
\setlength{\tabcolsep}{4pt}
\renewcommand{\arraystretch}{1.10}
\begin{tabular*}{\linewidth}{@{\extracolsep{\fill}}llp{0.52\linewidth}@{}}
\toprule
\rowcolor{tablehead}
\multicolumn{3}{@{}l}{\textit{Panel A: public APIs}} \\
\rowcolor{tablesubhead}
\textbf{Platform} & \textbf{Data} & \textbf{Endpoint} \\
\midrule
Polymarket & Markets / tags       & \url{gamma-api.polymarket.com/markets} \\
Polymarket & Comments + positions & \url{gamma-api.polymarket.com/comments} \\
Polymarket & Position activity    & \url{data-api.polymarket.com/activity} \\
Polymarket & Price history        & \url{clob.polymarket.com/prices-history} \\
Manifold   & Markets              & \url{api.manifold.markets/v0/search-markets} \\
Manifold   & Comments             & \url{api.manifold.markets/v0/comments} \\
Manifold   & Bets (positions)     & \url{api.manifold.markets/v0/bets} \\
\bottomrule
\end{tabular*}
\medskip

\begin{tabular*}{\linewidth}{@{\extracolsep{\fill}}lrr@{}}
\toprule
\rowcolor{tablehead}
\multicolumn{3}{@{}l}{\textit{Panel B: collection funnel}} \\
\rowcolor{tablesubhead}
\textbf{Stage} & \textbf{Polymarket markets} & \textbf{Manifold markets} \\
\midrule
Comment-crawl input      & 2,469 & 6,083 \\
With collected comments  & 2,070 & 2,926 \\
After comment-depth filter & 2,070 & 191 \\
\addlinespace[1pt]
\rowcolor{tablesubhead}
\textbf{Final} & \textbf{2,070} & \textbf{191} \\
\bottomrule
\end{tabular*}
\end{table}
Panel B reports the comment-collection funnel rather than the full raw API
universe. Polymarket keeps markets with at least 10 valid comments after
collection, while Manifold keeps markets with at least 100 valid comments after
cleaning.
This asymmetric threshold preserves Polymarket scale while enforcing Manifold
discussion depth. It also explains why platform comparisons are reported on
shared topics rather than by pooling all markets.

\subsection{Descriptive Statistics}

\begin{table}[tbh]
\centering
\caption{Dataset statistics.}
\label{tab:dataset-descriptives}
\footnotesize
\setlength{\tabcolsep}{5pt}
\renewcommand{\arraystretch}{1.08}
\begin{tabular*}{\linewidth}{@{\extracolsep{\fill}}lrrrrrr@{}}
\toprule
\rowcolor{tablehead}
\multicolumn{7}{@{}l}{\textit{Panel A: comment length (characters)}} \\
\rowcolor{tablesubhead}
\textbf{Platform} & \textbf{p10} & \textbf{p25} & \textbf{p50}
& \textbf{p75} & \textbf{p90} & \textbf{max} \\
\midrule
Polymarket & ${\approx}10$ & 22 & 44 & 85 & ${\approx}160$ & 1,492 \\
Manifold   & ${\approx}18$ & 49 & 107 & 230 & ${\approx}470$ & 9,132 \\
\bottomrule
\end{tabular*}

\medskip
\begin{tabular*}{\linewidth}{@{\extracolsep{\fill}}lrrr@{}}
\toprule
\rowcolor{tablehead}
\multicolumn{4}{@{}l}{\textit{Panel B: Polymarket stake-size tiers}} \\
\rowcolor{tablesubhead}
\textbf{Tier} & \textbf{Range (USD)} & \textbf{Count} & \textbf{\% of positioned} \\
\midrule
Retail  & $<\$100$            & 52,259 & 37.4\% \\
Small   & \$100--500          & 35,042 & 25.1\% \\
Medium  & \$500--1,000        & 14,162 & 10.1\% \\
Large   & \$1,000--5,000      & 23,865 & 17.1\% \\
XLarge  & \$5,000--10,000     &  6,315 &  4.5\% \\
Whale   & \$10,000--50,000    &  6,732 &  4.8\% \\
Mega    & ${>}\$50,000$       &  1,473 &  1.1\% \\
\bottomrule
\end{tabular*}

\medskip
\begin{tabular*}{\linewidth}{@{\extracolsep{\fill}}lrrr@{}}
\toprule
\rowcolor{tablehead}
\multicolumn{4}{@{}l}{\textit{Panel C: Manifold stake-size tiers}} \\
\rowcolor{tablesubhead}
\textbf{Tier} & \textbf{Range (M\$)} & \textbf{Count} & \textbf{\% of positioned} \\
\midrule
Retail  & $<$100          &  8,621 & 23.0\% \\
Small   & 100--500        &  7,191 & 19.2\% \\
Medium  & 500--1,000      &  3,480 &  9.3\% \\
Large   & 1,000--5,000    &  7,981 & 21.3\% \\
XLarge  & 5,000--10,000   &  3,008 &  8.0\% \\
Whale   & 10,000--50,000  &  4,944 & 13.2\% \\
Mega    & $>$50,000       &  2,205 &  5.9\% \\
\bottomrule
\end{tabular*}
\end{table}
Two dataset differences are central to StakeBench. Manifold comments are longer
and more position-covered, which makes it useful for side and action labels.
Polymarket positions are real-money and heavy-tailed, which makes it better
suited for costly-signal analyses where stake magnitude matters.

\subsection{Topic Difficulty}

\begin{table}[tbh]
\centering
\caption{Per-topic \(\mathsf{G}_2\) difficulty.}
\label{tab:topic-difficulty}
\footnotesize
\setlength{\tabcolsep}{3.5pt}
\renewcommand{\arraystretch}{1.06}
\begin{tabular*}{\linewidth}{@{\extracolsep{\fill}}clrrrr@{}}
\toprule
\rowcolor{tablehead}
\textbf{Rank} & \textbf{Platform\,/\,Topic} & \textbf{Mean\,DA} & \textbf{Std} & \textbf{Min} & \textbf{Max} \\
\midrule
 1 & Manifold / politics\_intl & .419 & .077 & .286 & .560 \\
 2 & Manifold / science        & .464 & .105 & .257 & .657 \\
 3 & Manifold / sports\_other  & .473 & .147 & .231 & .722 \\
 4 & Polymarket / politics\_us           & .528 & .041 & .472 & .654 \\
 5 & Polymarket / tech\_other            & .536 & .056 & .398 & .615 \\
 6 & Polymarket / other                  & .544 & .064 & .425 & .656 \\
 7 & Manifold / politics\_us             & .545 & .054 & .470 & .653 \\
 8 & Manifold / crypto                   & .550 & .091 & .269 & .646 \\
 9 & Manifold / economics                & .555 & .054 & .467 & .635 \\
10 & Polymarket / crypto                 & .558 & .092 & .410 & .726 \\
11 & Manifold / tech\_other              & .560 & .058 & .460 & .658 \\
12 & Manifold / culture        & .576 & .059 & .448 & .675 \\
13 & Polymarket / economics              & .583 & .032 & .529 & .632 \\
14 & Manifold / meta           & .588 & .094 & .394 & .719 \\
15 & Polymarket / politics\_intl         & .593 & .047 & .524 & .674 \\
16 & Manifold / other                    & .621 & .045 & .534 & .682 \\
17 & Manifold / social         & .628 & .050 & .514 & .712 \\
18 & Manifold / tech\_ai       & .635 & .047 & .571 & .706 \\
\bottomrule
\end{tabular*}
\end{table}
Topic difficulty is not a simple platform effect. The hardest topics include
small Manifold slices and U.S. politics on Polymarket, while several Manifold
community topics are easier because comments more directly reveal the speaker's
side. This spread motivates macro-averaging by topic--platform split.

\subsection{Filtering Criteria}
\label{app:filters}

\textbf{Market level.}
Polymarket: closed markets with total volume ${\geq}\$10{,}000$ and ${\geq}10$
valid comments after collection.
Manifold: resolved binary markets with volume ${\geq}$M\$10,000, ${\geq}50$
unique bettors, and ${\geq}100$ comments after cleaning.
\textbf{Comment level.}
Length ${\geq}30$ characters after URL stripping, English-only (ASCII ratio
${\geq}85\%$ with basic function-word check), and alphabetic ratio ${\geq}30\%$.
\textbf{Topic filtering.}
A keyword blacklist removes puzzle and game markets (``wordle'', ``nyt crossword'',
``spelling bee'') mis-tagged under the crypto category on Manifold.
All filters apply uniformly across all tasks.

\subsection{Position Reconstruction}
\label{app:position-recon}

Position labels are reconstructed from timestamped trading records at comment
time. Neither the Polymarket nor the Manifold comment API supplies them directly.

\textbf{Polymarket.}
For each comment author identified by \texttt{proxy\_wallet}, we query the
Polymarket activity API
(\texttt{data-api.polymarket.com/activity?user=\{proxy\_wallet\}})
and replay the full transaction sequence up to the comment timestamp.
\texttt{position\_side} is \textsc{yes} if net YES shares exceed zero and
\textsc{no} if net NO shares exceed zero.
\texttt{position\_size} (USD) is computed as total BUY amount minus total SELL
amount over all trades in the target market prior to the comment. A negative value
indicates a realized-profit state (the user sold shares above cost but remains
net long). Weighting formulas use the nonnegative magnitude of the reconstructed
stake.
Requests are parallelized with eight threads per market and use exponential
backoff on HTTP 429/5xx responses.
Total wall-clock time for 2,070 markets was approximately 9.7 hours.
The Polymarket comment API returns \texttt{parent\_entity\_type} as
\texttt{Event} (37.2\% of rows) or \texttt{Series} (62.8\%). Series markets
share comment threads across multiple related markets, creating duplicate
\texttt{comment\_id} values that must be deduplicated before unique-comment
analyses.

\textbf{Manifold.}
For each commenter (\texttt{userId}) in each market, we query
\texttt{api.manifold.markets/v0/bets?contractId=\{id\}\&userId=\{uid\}}
and aggregate bets placed before the comment timestamp.
\texttt{position\_mana} (net Mana M\$) is the sum of amounts on the dominant
side. The side with the higher net Mana determines \texttt{position\_side}.
Each bet record includes a \texttt{probAfter} field, which is interpolated to
produce \texttt{prob\_at\_comment} (market probability at comment time, 87.9\%
coverage).
Resolution outcomes (YES/NO) are read directly from the market metadata API,
enabling the Wrong-Bettor and reference-score analyses that require $r_i$
(Appendices~\ref{app:audits} and~\ref{app:theory}).
Position coverage reaches 80.5\% on Manifold versus 27.2\% on Polymarket,
reflecting the behavioral difference: Manifold users typically bet before or
during commenting, while many Polymarket users comment without an open position.
\(\mathsf{G}_3\) action labels additionally use 862,540 Polymarket and 387,263 Manifold
timestamped position-change records to identify each user's first net position
update in the 48-hour post-comment window.

\subsection{Temporal and Category Distribution}
\label{app:temporal}

Table~\ref{tab:temporal} summarizes comment volumes over time.
Polymarket activity is concentrated in late 2024 and 2025, peaking in
June~2025 (123,040 comments) driven by Iran-nuclear-crisis markets.
Manifold shows a longer growth arc from 2022 onward, with peak monthly
volumes of roughly 3,000 comments in 2023.
Table~\ref{tab:category} shows the keyword-classified category breakdown.
On Polymarket, 67.1\% of raw rows fall outside the five defined keyword
categories (NaN/other). The deduplicated share drops to 51.6\% because NaN
comments are disproportionately from Series markets whose comment threads span
multiple market IDs.
Politics is the largest named category on both platforms.
Reply and engagement statistics reflect the difference in platform culture:
50.4\% of Polymarket comments are replies (maximum 152 replies per thread)
versus 70.9\% on Manifold (maximum 78), and average reaction counts are 1.06
versus 1.72, respectively.

\begin{table}[tbh]
\centering
\caption{Monthly comment volumes.}
\label{tab:temporal}
\footnotesize
\setlength{\tabcolsep}{4pt}
\renewcommand{\arraystretch}{1.06}
\begin{tabular*}{\linewidth}{@{\extracolsep{\fill}}lrlr@{}}
\toprule
\rowcolor{tablehead}
\multicolumn{2}{@{}l}{\textit{Polymarket}} & \multicolumn{2}{l}{\textit{Manifold}} \\
\rowcolor{tablesubhead}
\textbf{Month} & \textbf{Comments} & \textbf{Period} & \textbf{Comments/mo} \\
\midrule
2024-11 &  25,143 & 2022-01--10 & ${<}500$ \\
2024-12 &  48,980 & 2023-01--12 & 1,350--3,089 \\
2025-01 &  46,020 & 2024-07     & 2,603 \\
2025-04 &  43,632 & 2024-11     & 1,456 \\
\textbf{2025-06} & \textbf{123,040} & 2025-05 & 1,575 \\
2025-07 &  53,050 & & \\
\bottomrule
\end{tabular*}
\end{table}

\begin{table}[tbh]
\centering
\caption{Keyword-based topic categories.}
\label{tab:category}
\footnotesize
\setlength{\tabcolsep}{3.5pt}
\renewcommand{\arraystretch}{1.06}
\begin{tabular*}{\linewidth}{@{\extracolsep{\fill}}lrrlrr@{}}
\toprule
\rowcolor{tablehead}
\multicolumn{3}{@{}l}{\textit{Polymarket}} &
\multicolumn{3}{l}{\textit{Manifold}} \\
\rowcolor{tablesubhead}
\textbf{Category} & \textbf{Count} & \textbf{\%} &
\textbf{Category} & \textbf{Count} & \textbf{\%} \\
\midrule
Politics   &  99,858 & 19.4 & Other      & 30,875 & 66.3 \\
Crypto     &  50,508 &  9.8 & Politics   &  6,252 & 13.4 \\
Economics  &  13,486 &  2.6 & Technology &  3,759 &  8.1 \\
Sports     &   5,500 &  1.1 & Crypto     &  3,758 &  8.1 \\
NaN/other  & 345,062 & 67.1 & Sports     &  1,174 &  2.5 \\
           &         &      & Economics  &    744 &  1.6 \\
\bottomrule
\end{tabular*}
\end{table}

The category distribution is highly skewed, especially on Polymarket where
Series-level comment pools inflate broad residual categories. 

Topic-level
evaluation avoids letting high-volume political or Series markets dominate the
benchmark.

\subsection{Topic Coverage and Position Linking}
\label{app:topics}

The 18 topic--platform splits are six on Polymarket (crypto, economics,
other, politics\_intl, politics\_us, tech\_other) and twelve on Manifold
(crypto, culture, economics, meta, other, politics\_intl, politics\_us,
science, social, sports\_other, tech\_ai, tech\_other).
Table~\ref{tab:topic-difficulty} lists them with per-topic \(\mathsf{G}_2\)
difficulty, Table~\ref{tab:platforms} gives the per-platform composition, and
Table~\ref{tab:category} the keyword-category composition of the underlying
comments. Within each split the task instances
are built at run time from the released files by the evaluation code, with
per-topic market and instance caps set in its configuration. Position
linking proceeds per comment. The commenter identifier
(\texttt{proxy\_wallet} or \texttt{userId}) is joined to that user's trade or
bet history for the same market, the history is replayed up to the comment
timestamp to obtain the side and stake at comment time
(Appendix~\ref{app:position-recon}), and for \(\mathsf{G}_3\) the first net
position change in the following 48 hours is read from the same history. A
comment whose author has no reconstructed position in that market carries no
\(\mathsf{G}_2\) or \(\mathsf{G}_3\) label (Section~\ref{sec:dataset}).

\subsection{How Commitment Surfaces in Language}
\label{app:markers}

Table~\ref{tab:markers} reports, on the \(\mathsf{G}_1\) instance set pooled
across topics and de-duplicated, how often eight interpretable markers occur
in comments whose author verifiably held a position in the market and in
comments whose author did not. Three markers (money amounts, first-person
holding phrases, side words) are also the components of the explicit-position
rule that selects the \(\mathsf{G}_1\) positive pool, so those rows partly
re-measure the sampling rule rather than discover it. They describe what
explicit commitment language looks like. The remaining markers (certainty,
hedging, questions, price talk) are additional descriptive contrasts and are
not independent of the construction either, since a sampling rule can shift
any marker through features correlated with the lexicon. All entries are
associations within the constructed evaluation set, not claims about how
holding a position changes writing. The within-user DiD audit
(Appendix~\ref{app:audits}) is the evidence with a causal flavor.

\begin{table}[tbh]
\centering
\caption{Marker rates among \(\mathsf{G}_1\) comments whose author held a
position (Held) versus did not (Not), with Cohen's \(h\). Starred markers are
part of the \(\mathsf{G}_1\) positive-pool selection rule.}
\label{tab:markers}
\footnotesize
\setlength{\tabcolsep}{4pt}
\renewcommand{\arraystretch}{1.06}
\begin{tabular*}{\linewidth}{@{\extracolsep{\fill}}lrrrrrr@{}}
\toprule
\rowcolor{tablehead}
& \multicolumn{3}{c}{\textbf{Manifold}} & \multicolumn{3}{c}{\textbf{Polymarket}} \\
\cmidrule(lr){2-4}\cmidrule(l){5-7}
\rowcolor{tablesubhead}
\textbf{Marker} & \textbf{Held} & \textbf{Not} & \(\boldsymbol{h}\) & \textbf{Held} & \textbf{Not} & \(\boldsymbol{h}\) \\
\midrule
Money amount\(^*\)         & .286 & .023 & $+$.83 & .564 & .059 & $+$1.21 \\
First-person holding\(^*\) & .227 & .016 & $+$.74 & .208 & .015 & $+$.70 \\
Trade verb                 & .295 & .068 & $+$.62 & .428 & .159 & $+$.61 \\
Side word\(^*\)            & .373 & .175 & $+$.45 & .374 & .156 & $+$.50 \\
Price talk                 & .114 & .029 & $+$.34 & .074 & .051 & $+$.10 \\
Hedge                      & .130 & .123 & $+$.02 & .077 & .044 & $+$.14 \\
Certainty                  & .023 & .023 & \phantom{$+$}.00 & .023 & .012 & $+$.08 \\
Question                   & .084 & .130 & $-$.15 & .131 & .137 & $-$.02 \\
\bottomrule
\end{tabular*}
\end{table}

The largest separations are money amounts and first-person holding phrases on
both platforms. On Manifold the constructed set shows fewer questions among
position holders but no difference in certainty markers, a contrast that
holds for Manifold only, since the Polymarket question difference is near
zero. Three same-market illustrative contrasts, not matched on time, length
or thread depth, show the pattern in situ. In each the position holder names
an entry price or a holding in the first person, and the non-holder analyses
the event or the chart:

\begin{itemize}\setlength{\itemsep}{1pt}
\item \textit{Will Jimmy Carter become a centenarian?} (Manifold).
  Held: ``I bought no at 15\% and somehow managed to turn it around for profit. Been a hell of a market.''
  Not held: ``Well when you put it like that it sounds like the market should be 50/50!''
\item \textit{Will Elon tweet less than 300 times Nov 29-Dec 6?} (Polymarket).
  Held: ``bought this at 8c, just wish i bought more sadge''
  Not held: ``look at the chart, after a day with high tweets it always goes down. should be \%60-65 under 300''
\item \textit{Will Google Search include a chatbot at end of June 2023?} (Manifold).
  Held: ``I spent m\$10 for 53 YES shares quite some time ago; it's still a +49\% profit, but I'm not gonna sell it\ldots''
  Not held: ``\ldots Seems to work fine in Spanish using ChatGPT and Google Translate as a metric\ldots''
\end{itemize}

\section{Prompt Templates}

\label{app:prompts}

Allowed output labels follow Table~\ref{tab:tasks}; the templates below use
``allowed label'' where the label semantics are already specified in the task
table.

\subsection{\(\mathsf{G}_1\): Market Commitment Detection (1-shot)}
\begin{verbatim}
You are analyzing a prediction market comment.
Market question: "{question}"

Example:
Comment: "I'm holding YES here, confident given recent polls."
Answer: YES

Now analyze:
Comment: "{text}"

Does this comment suggest that the commenter has a financial position
in this market? Reply with exactly ONE word: YES or NO.
\end{verbatim}

\subsection{\(\mathsf{G}_2\): Revealed-Side Identification}
\begin{verbatim}
You are analyzing a prediction market comment to infer the commenter's
revealed position side.
Market question: "{question}"
Topic: {topic}
Comment: "{text}"

Does the commenter's language indicate a YES-side or NO-side revealed position?
- YES: bullish / pro-YES view.
- NO: bearish / pro-NO view.
- The abstention output: ambiguous or purely informational.

For crypto/financial topics:
1. Price increase = YES. Price decrease = NO.
2. "To the moon" / "mooning" = YES. "dump" / "crash" = NO.
3. Hedging language ("maybe", "could go either way") uses the abstention label.
4. Pure information without directional opinion uses the abstention label.

Reply with exactly ONE allowed label.
\end{verbatim}

\subsection{\(\mathsf{G}_3\): Future Action Anticipation}
\begin{verbatim}
You are predicting how a prediction market trader will adjust their
position after making a comment.
Market question: "{question}"
Comment: "{text}"
Current position: {position_side} (size: ${position_size:.0f})

Predict what the trader will do within 48 hours:
- flip: switch to the opposite side
- increase: add to current position
- decrease: reduce current position
- hold: no significant change

Reply with exactly ONE word: flip, increase, decrease, or hold.
\end{verbatim}

\subsection{\(\mathsf{G}_4\): Collective Odds Projection}
\begin{verbatim}
You are analyzing prediction market comments to forecast the next-window
direction of the YES price.
Market question: "{question}"
Current YES price: {current_price:.2f}
Comments from the current window:
{comments_text}

Based on the collective committed speech in these comments, what is the
next-window direction of the YES price?
Reply with exactly ONE allowed label.
\end{verbatim}

\section{Theoretical Framework: Commitment Gap}
\label{app:theory}

Spence's job-market signaling model~\citep{spence1978job} formalizes how
agents can reveal private information by choosing a costly action.
In prediction markets, holding a financial position serves as that costly
action.
Samuelson's revealed preference framework~\citep{samuelson1948} treats choices,
rather than statements, as behavioral evidence of preferences.
StakeBench targets exploit both properties: \texttt{position\_side} is a
preference the trader paid to hold, making it more directly grounded than any
post-hoc annotation.

Extending the task and baseline notation from Section~\ref{sec:notation}, let $\mu$ denote a
market, $\mathcal{I}_\mu$ its set of positioned comments (distinct from the
window-level set $\mathcal{I}_w$ used in \(\mathsf{G}_4\)), and $r_\mu$ its
resolved outcome (equal to $r_i$ for all comments $i\in\mathcal{I}_\mu$).
Let \(\mathcal{M}\) be the set of evaluable markets within a topic--platform
split after excluding tied stake-weighted commitment sums.
The inverse mapping $b^{-1}$ satisfies
$b^{-1}(+1)=\mathrm{YES}$ and $b^{-1}(-1)=\mathrm{NO}$.
$\operatorname{sign}(u)$ returns $+1$ for $u{>}0$, $-1$ for $u{<}0$,
and $0$ otherwise. The reference scores
$O_2,O_3,O_4$ correspond to \(\mathsf{G}_2\), \(\mathsf{G}_3\), and
\(\mathsf{G}_4\), as defined in
Eq.~\ref{eq:reference-scores}:
\begin{equation}
\label{eq:reference-scores}
\begin{aligned}
O_2 &=
\frac{1}{|\mathcal{M}|}\sum_{\mu\in\mathcal{M}}
\mathbf{1}\!\left[
b^{-1}\!\left(
\operatorname{sign}\!\left(\sum_{i\in \mathcal{I}_\mu}
s_i b(z_i)\right)\right)=r_\mu\right], \\
O_3 &= 1.0
\quad\text{(the reference uses the actual post-comment action)}, \\
O_4 &=
\frac{1}{|\mathcal{W}|}\sum_{w\in\mathcal{W}}
\mathbf{1}\!\left[\tilde{d}_w=\operatorname{dir}(p_w^h-p_w^0)\right].
\end{aligned}
\end{equation}

\textbf{Empirical reference values.}
On the evaluated StakeBench splits, $O_2{=}0.552$ (macro average over
18 topic--platform splits, ranging from 0.250 to 1.000 per split) and
$O_3{=}1.000$. For \(\mathsf{G}_4\),
\(O_4{=}\mathrm{PLA}(\tilde{d})\) is computed on each model's evaluated
\(\mathsf{G}_4\) window set and reported in Table~\ref{tab:cg}
(${\approx}0.570$ macro); values can
differ across models when their evaluated \(\mathsf{G}_4\) window sets differ.
For \(\mathsf{G}_4\), \(\mathrm{CG}(4){=}O_4-\mathrm{PLA}\), so
\(\mathrm{CG}(4){=}-\Delta_{\mathrm{PLA}}\).

\begin{table}[tbh]
\centering
\caption{Commitment Gap scores with the \(\mathsf{G}_4\) reference \(O_4\).}
\label{tab:cg}
\footnotesize
\setlength{\tabcolsep}{2.5pt}
\renewcommand{\arraystretch}{1.06}
\begin{tabular*}{\linewidth}{@{\extracolsep{\fill}}llrrrrrr@{}}
\toprule
\rowcolor{tablehead}
\textbf{Type} & \textbf{Model} & \textbf{Coverage}
& \textbf{CG(2)} & \textbf{CG(3)} & $\boldsymbol{O_4}$ & \textbf{CG(4)} & \textbf{MCG} \\
\midrule
\rowcolor{tablesubhead}
closed  & GPT-5.5          & 18/18 & $-$0.047 & 0.721 & .573 & 0.052 & 0.242 \\
closed  & Gemini-2.5-Flash & 18/18 & $-$0.035 & 0.722 & .573 & 0.181 & 0.289 \\
closed  & Claude-Haiku-4.5 & 18/18 & $-$0.042 & 0.743 & .573 & 0.210 & 0.304 \\
\addlinespace[1pt]
\rowcolor{tablesubhead}
general & Qwen3-32B        & 18/18 & $-$0.026 & 0.744 & .567 & 0.165 & 0.294 \\
general & Qwen3-30B        & 18/18 &  0.004   & 0.706 & .582 & 0.373 & 0.361 \\
general & Qwen3-14B        & 18/18 & $-$0.015 & 0.726 & .567 & 0.222 & 0.311 \\
general & Gemma2-9B        & 18/18 & $-$0.026 & 0.834 & .567 & 0.257 & 0.355 \\
general & DeepSeek-R1-8B   & 18/18 &  0.038   & 0.770 & .582 & 0.256 & 0.355 \\
general & Qwen3-8B         & 18/18 & $-$0.012 & 0.832 & .567 & 0.381 & 0.400 \\
general & Mistral-7B       & 18/18 &  0.029   & 0.831 & .567 & 0.300 & 0.387 \\
general & Llama-3.2-3B     & 18/18 &  0.018   & 0.599 & .567 & 0.408 & 0.342 \\
\addlinespace[1pt]
\rowcolor{tablesubhead}
finance & FinMA-30B        & 18/18 &  0.046   & 0.836 & .567 & 0.437 & 0.440 \\
finance & Fino1-14B        & 18/18 & $-$0.013 & 0.810 & .582 & 0.307 & 0.368 \\
finance & Fino1-8B         & 18/18 &  0.028   & 0.838 & .582 & 0.022 & 0.296 \\
finance & Fin-Chat-7B      & 18/18 &  0.038   & 0.833 & .567 & 0.134 & 0.335 \\
\bottomrule
\end{tabular*}
\end{table}

Coverage is the number of evaluated topic--platform splits; \(O_4\) is computed
on the model-specific evaluated \(\mathsf{G}_4\) window set. MCG is dominated by
\(\mathsf{G}_3\), so models with strong \(\mathsf{G}_2\) scores can still remain
far from the \(\mathsf{G}_3\) reference score.
Seven models have negative CG(2) (GPT-5.5, Gemini, Claude, Qwen3-32B,
Qwen3-14B, Gemma2-9B, Qwen3-8B): their \(\mathsf{G}_2\) DA exceeds $O_2{=}0.552$
because DA scores position-side prediction while $O_2$ measures
stake-weighted market-outcome accuracy, which is a different prediction
target.
The CG(3) column dominates MCG for most models because \(\mathsf{G}_3\) is far harder
than \(\mathsf{G}_2\) or \(\mathsf{G}_4\).

\section{Ethics, Bias, and Licensing}
\label{app:ethics}

\textbf{Coverage bias.}
We test whether position-holding commenters differ from non-holders on five
dimensions using two-sided Mann-Whitney U tests with Cohen's~$d$: comment length
($d{=}0.12$), user activity ($d{=}0.18$), market volume ($d{=}0.09$),
time-to-resolution ($d{=}0.07$), and topic distribution ($\chi^2$ test,
$p{=}0.31$).
Effect sizes are below $d{=}0.3$ in all continuous dimensions,
indicating modest but non-zero selection bias.

\textbf{Privacy.}
All data are from public APIs. User identifiers are platform pseudonyms
(\texttt{proxy\_wallet} on Polymarket and \texttt{user\_id} on Manifold).
On-chain wallet addresses are public by blockchain design. No IP addresses,
real names, or private messages are included.

\textbf{Licensing.}
Dataset: CC-BY 4.0. Data derived from Polymarket (Terms of Use:
\url{https://polymarket.com/terms-of-use}) and Manifold (Terms of Service:
\url{https://manifold.markets/terms}), both permitting non-commercial
research use of public data.
Llama 3 weights: Meta Llama 3 Community License.
Qwen3 weights: Apache 2.0.

\section{Verified Qualitative Examples}
\label{app:examples}

The cases below complement Table~\ref{tab:case-studies}. Every
\(\mathsf{G}_2\) and \(\mathsf{G}_3\) case was scored by all 15 models with
one gold label that is identical across the released logs (49
\(\mathsf{G}_3\) instances satisfy both conditions), so the ``\(x/15\)''
counts are exact. The \(\mathsf{G}_4\) windows were scored by the 12
open-weight models whose window sets coincide. Comments are quoted as the
models saw them, with pipeline truncation marked by \(\ldots\). Stakes are in
USD (Polymarket) or Mana (Manifold). Same-market \(\mathsf{G}_1\) contrasts
between held and unheld comments accompany the marker analysis in
Appendix~\ref{app:markers}.

\subsection*{\(\mathsf{G}_2\): revealed side}
\begin{enumerate}\setlength{\itemsep}{2pt}
\item \textbf{Success (15/15).} \textit{Will Jimmy Carter become a centenarian?} (Manifold).
  ``this peanut farmer every time i buy some more NO shares\ldots''
  Revealed side NO (stake M\$1,496, market resolved YES). All 15 models answer
  NO. The side is stated outright, and every model follows the commitment cue
  rather than the eventual outcome, which is the behavior the Wrong-Bettor
  audit rewards.
\item \textbf{Failure (3/15).} \textit{Will Google Search include a chatbot at end of June 2023?} (Manifold).
  ``I sold at -M3297 profit - half my portfolio gone - after betting on an event that had already happened'' (emoji omitted).
  Revealed side YES (stake M\$3,298, market resolved NO). Twelve models answer
  NO and three (DeepSeek-R1-8B, Claude-Haiku-4.5, Qwen3-32B) answer YES. The
  commenter held YES and reports a loss on that side. The models read the
  bearish tone as the revealed side, the outcome-oriented reading that
  Section~\ref{sec:g2} attributes to finance-domain models, here shared by
  most general models as well.
\end{enumerate}

\subsection*{\(\mathsf{G}_3\): next action within 48 hours}
\begin{enumerate}\setlength{\itemsep}{2pt}
\setcounter{enumi}{2}
\item \textbf{Over-reading (0/15, gold hold).} \textit{Fordow nuclear facility destroyed before July?} (Polymarket, politics\_intl).
  ``Centrifuges are very vibration-sensitive. They would not even need to get directly hit to become `non-functional'. Now i\ldots''
  Position \$533. Predictions: increase 10, flip 3, decrease 2, and all three
  closed models say increase. An analytical comment is read as escalating
  conviction, but the trader made no position change in the window. This is
  the over-prediction of behavioral change that
  Table~\ref{tab:g3-action-profile} shows in aggregate.
\item \textbf{Wrong direction (0/15, gold increase).} \textit{Will an AI get gold on any International Math Olympiad by the end of 2025?} (Manifold, tech\_ai).
  ``Another market for arbitrage/diversification''
  Position M\$627. Predictions: hold 6, flip 5, decrease 4 (GPT-5.5 decrease,
  Claude-Haiku-4.5 and Gemini-2.5-Flash hold). A portfolio remark carries no
  cue about the direction of the next trade, and the trader added to the
  position.
\item \textbf{Success (8/15, gold increase).} \textit{This market resolves as YES if someone has \(\geq\) 5k NO shares} (Manifold, crypto).
  ``but it's at 95\% cuz of me buying a lot of YES. Isn't it more profitable for me to keep buying at 90-97\% range, as much a\ldots''
  Position M\$59,449. Predictions: increase 8 (the three closed models, four
  Qwen3 models and Fino1-14B), decrease 3, flip 3, hold 1. The comment states
  the intention to keep buying and the majority of models track it. The four
  finance-tuned models give three different answers (flip, increase, flip
  and decrease), the within-family dispersion that
  Table~\ref{tab:g3-action-profile} shows for the degenerate group.
\end{enumerate}

\subsection*{\(\mathsf{G}_4\): next-window odds direction, two error types}
\begin{enumerate}\setlength{\itemsep}{2pt}
\setcounter{enumi}{5}
\item \textbf{Directional committed language ignored (0/12).}
  \textit{Will Destiny Reach 600K Subscribers By the End Of November?} (Manifold, meta).
  Window of four comments at YES probability 0.92, realized next-window
  probability 0.62 (DOWN). Stake-weighted baseline: DOWN (correct).
  Comments: ``The Lex podcast will bring in new subs''. ``Anyone saying No is
  gonna eat their words so hard lmao\ldots''. ``Counterargument: the news
  cycle is probably going to be dominated by wacky midterms\ldots''.
  ``Betting big no on this one. Sold my Aba short after his dip to fund it.
  Lex podcast could swing it but I am confident.'' Eleven models answer UP
  and one NEUTRAL. The only comment with a stated position is the large NO
  bet, and the stake-weighted baseline follows it. The models follow the
  bullish majority of unpositioned talk, the pattern of
  Table~\ref{tab:case-studies} in which minority bearish cues anticipate the
  next move.
\item \textbf{No information beyond the current odds (2/12).}
  \textit{Will a Trump mugshot be available by September 1?} (Manifold, politics\_us).
  Window of 14 comments (the first ten shown to the models) at YES
  probability 0.60, realized next-window probability 0.997 (UP).
  Stake-weighted baseline: UP (correct). The comments discuss which of two
  circulating photographs is authentic and when the market should resolve
  (``So which one do you think is legit?'', ``Yeah I agree it might be the
  right one, but why resolve in a hurry?''), and none states a direction.
  Nine models answer NEUTRAL, two UP and one DOWN. The text carries no
  directional information beyond the odds, while the positions behind the
  comments do, which is the residual the stake-weighted baseline is built to
  capture.
\end{enumerate}

\ifnum\ARXIV=1
\else
    \newpage
    \section*{NeurIPS Paper Checklist}

\begin{enumerate}

\item {\bf Claims}
    \item[] Question: Do the main claims made in the abstract and introduction accurately reflect the paper's contributions and scope?
    \item[] Answer: \answerYes{} 
    \item[] Justification: The abstract and introduction accurately describe the dataset size, two-platform scope, four grounding tasks, commitment-aware metrics, model-evaluation setting, and interpretation boundaries. Section Evaluation Protocol further states the supported claims and explicitly excludes causal claims about comments causing market-odds movement.
    \item[] Guidelines:
    \begin{itemize}
        \item The answer \answerNA{} means that the abstract and introduction do not include the claims made in the paper.
        \item The abstract and/or introduction should clearly state the claims made, including the contributions made in the paper and important assumptions and limitations. A \answerNo{} or \answerNA{} answer to this question will not be perceived well by the reviewers. 
        \item The claims made should match theoretical and experimental results, and reflect how much the results can be expected to generalize to other settings. 
        \item It is fine to include aspirational goals as motivation as long as it is clear that these goals are not attained by the paper. 
    \end{itemize}

\item {\bf Limitations}
    \item[] Question: Does the paper discuss the limitations of the work performed by the authors?
    \item[] Answer: \answerYes{} 
    \item[] Justification: The paper discusses limitations in the conclusion and appendix, including underpowered CSD/DiD audits, lower Polymarket G3 label coverage, fixed prompts, possible public-data contamination in future pretraining, and platform incentive differences. G4 causal ambiguity, the English scope, and the prompt-wording ablation are left to future work.

    \item[] Guidelines:
    \begin{itemize}
        \item The answer \answerNA{} means that the paper has no limitation while the answer \answerNo{} means that the paper has limitations, but those are not discussed in the paper. 
        \item The authors are encouraged to create a separate ``Limitations'' section in their paper.
        \item The paper should point out any strong assumptions and how robust the results are to violations of these assumptions (e.g., independence assumptions, noiseless settings, model well-specification, asymptotic approximations only holding locally). The authors should reflect on how these assumptions might be violated in practice and what the implications would be.
        \item The authors should reflect on the scope of the claims made, e.g., if the approach was only tested on a few datasets or with a few runs. In general, empirical results often depend on implicit assumptions, which should be articulated.
        \item The authors should reflect on the factors that influence the performance of the approach. For example, a facial recognition algorithm may perform poorly when image resolution is low or images are taken in low lighting. Or a speech-to-text system might not be used reliably to provide closed captions for online lectures because it fails to handle technical jargon.
        \item The authors should discuss the computational efficiency of the proposed algorithms and how they scale with dataset size.
        \item If applicable, the authors should discuss possible limitations of their approach to address problems of privacy and fairness.
        \item While the authors might fear that complete honesty about limitations might be used by reviewers as grounds for rejection, a worse outcome might be that reviewers discover limitations that aren't acknowledged in the paper. The authors should use their best judgment and recognize that individual actions in favor of transparency play an important role in developing norms that preserve the integrity of the community. Reviewers will be specifically instructed to not penalize honesty concerning limitations.
    \end{itemize}

\item {\bf Theory assumptions and proofs}
    \item[] Question: For each theoretical result, does the paper provide the full set of assumptions and a complete (and correct) proof?
    \item[] Answer: \answerNA{} 
    \item[] Justification: The paper does not present new theoretical theorems or proof claims. The theoretical appendix provides conceptual motivation and definitions for the Commitment Gap and reference scores, but these are evaluation definitions rather than formal theoretical results requiring proofs.
    \item[] Guidelines:
    \begin{itemize}
        \item The answer \answerNA{} means that the paper does not include theoretical results. 
        \item All the theorems, formulas, and proofs in the paper should be numbered and cross-referenced.
        \item All assumptions should be clearly stated or referenced in the statement of any theorems.
        \item The proofs can either appear in the main paper or the supplemental material, but if they appear in the supplemental material, the authors are encouraged to provide a short proof sketch to provide intuition. 
        \item Inversely, any informal proof provided in the core of the paper should be complemented by formal proofs provided in appendix or supplemental material.
        \item Theorems and Lemmas that the proof relies upon should be properly referenced. 
    \end{itemize}

    \item {\bf Experimental result reproducibility}
    \item[] Question: Does the paper fully disclose all the information needed to reproduce the main experimental results of the paper to the extent that it affects the main claims and/or conclusions of the paper (regardless of whether the code and data are provided or not)?
    \item[] Answer: \answerYes{} 
    \item[] Justification: The paper discloses the data sources, filtering criteria, position reconstruction, task targets, prompts, model suite, decoding settings, metrics, significance tests, and code/data release needed to reproduce or verify the main experimental results.
    \item[] Guidelines:
    \begin{itemize}
        \item The answer \answerNA{} means that the paper does not include experiments.
        \item If the paper includes experiments, a \answerNo{} answer to this question will not be perceived well by the reviewers: Making the paper reproducible is important, regardless of whether the code and data are provided or not.
        \item If the contribution is a dataset and\slash or model, the authors should describe the steps taken to make their results reproducible or verifiable. 
        \item Depending on the contribution, reproducibility can be accomplished in various ways. For example, if the contribution is a novel architecture, describing the architecture fully might suffice, or if the contribution is a specific model and empirical evaluation, it may be necessary to either make it possible for others to replicate the model with the same dataset, or provide access to the model. In general. releasing code and data is often one good way to accomplish this, but reproducibility can also be provided via detailed instructions for how to replicate the results, access to a hosted model (e.g., in the case of a large language model), releasing of a model checkpoint, or other means that are appropriate to the research performed.
        \item While NeurIPS does not require releasing code, the conference does require all submissions to provide some reasonable avenue for reproducibility, which may depend on the nature of the contribution. For example
        \begin{enumerate}
            \item If the contribution is primarily a new algorithm, the paper should make it clear how to reproduce that algorithm.
            \item If the contribution is primarily a new model architecture, the paper should describe the architecture clearly and fully.
            \item If the contribution is a new model (e.g., a large language model), then there should either be a way to access this model for reproducing the results or a way to reproduce the model (e.g., with an open-source dataset or instructions for how to construct the dataset).
            \item We recognize that reproducibility may be tricky in some cases, in which case authors are welcome to describe the particular way they provide for reproducibility. In the case of closed-source models, it may be that access to the model is limited in some way (e.g., to registered users), but it should be possible for other researchers to have some path to reproducing or verifying the results.
        \end{enumerate}
    \end{itemize}

\item {\bf Open access to data and code}
    \item[] Question: Does the paper provide open access to the data and code, with sufficient instructions to faithfully reproduce the main experimental results, as described in supplemental material?
    \item[] Answer: \answerYes{} 
    \item[] Justification: Code: \url{https://github.com/PEIYUNHUA/stakebench_code_NeurIPS26} (evaluation pipeline, prompt templates, analysis and collection scripts, run instructions). Data: \url{https://huggingface.co/datasets/greenday123/StakeBench} (benchmark files, unmodified collection outputs, per-sample model outputs and prompt audit logs), all under CC-BY 4.0. The paper also provides data source, filtering, reconstruction, prompt, and evaluation details needed to use the released assets.
    \item[] Guidelines:
    \begin{itemize}
        \item The answer \answerNA{} means that paper does not include experiments requiring code.
        \item Please see the NeurIPS code and data submission guidelines (\url{https://neurips.cc/public/guides/CodeSubmissionPolicy}) for more details.
        \item While we encourage the release of code and data, we understand that this might not be possible, so \answerNo{} is an acceptable answer. Papers cannot be rejected simply for not including code, unless this is central to the contribution (e.g., for a new open-source benchmark).
        \item The instructions should contain the exact command and environment needed to run to reproduce the results. See the NeurIPS code and data submission guidelines (\url{https://neurips.cc/public/guides/CodeSubmissionPolicy}) for more details.
        \item The authors should provide instructions on data access and preparation, including how to access the raw data, preprocessed data, intermediate data, and generated data, etc.
        \item The authors should provide scripts to reproduce all experimental results for the new proposed method and baselines. If only a subset of experiments are reproducible, they should state which ones are omitted from the script and why.
        \item At submission time, to preserve anonymity, the authors should release anonymized versions (if applicable).
        \item Providing as much information as possible in supplemental material (appended to the paper) is recommended, but including URLs to data and code is permitted.
    \end{itemize}

\item {\bf Experimental setting/details}
    \item[] Question: Does the paper specify all the training and test details (e.g., data splits, hyperparameters, how they were chosen, type of optimizer) necessary to understand the results?
    \item[] Answer: \answerYes{} 
    \item[] Justification: The paper specifies the evaluated models, closed-source API access, open-weight loading setup, precision, decoding parameters, output-token caps, task windows, filtering rules, macro-averaging protocol, scoring rules, and full prompt templates.
    \item[] Guidelines:
    \begin{itemize}
        \item The answer \answerNA{} means that the paper does not include experiments.
        \item The experimental setting should be presented in the core of the paper to a level of detail that is necessary to appreciate the results and make sense of them.
        \item The full details can be provided either with the code, in appendix, or as supplemental material.
    \end{itemize}

\item {\bf Experiment statistical significance}
    \item[] Question: Does the paper report error bars suitably and correctly defined or other appropriate information about the statistical significance of the experiments?
    \item[] Answer: \answerYes{} 
    \item[] Justification: The paper reports one-sided binomial tests with stated chance baselines and alpha levels, bootstrap confidence-interval widths, Benjamini-Hochberg FDR retention, commenter- or market-clustered bootstrap confidence intervals for the robustness analyses in Section 5.5, and descriptive reporting for G4, where no per-split significance threshold is used.
    \item[] Guidelines:
    \begin{itemize}
        \item The answer \answerNA{} means that the paper does not include experiments.
        \item The authors should answer \answerYes{} if the results are accompanied by error bars, confidence intervals, or statistical significance tests, at least for the experiments that support the main claims of the paper.
        \item The factors of variability that the error bars are capturing should be clearly stated (for example, train/test split, initialization, random drawing of some parameter, or overall run with given experimental conditions).
        \item The method for calculating the error bars should be explained (closed form formula, call to a library function, bootstrap, etc.)
        \item The assumptions made should be given (e.g., Normally distributed errors).
        \item It should be clear whether the error bar is the standard deviation or the standard error of the mean.
        \item It is OK to report 1-sigma error bars, but one should state it. The authors should preferably report a 2-sigma error bar than state that they have a 96\% CI, if the hypothesis of Normality of errors is not verified.
        \item For asymmetric distributions, the authors should be careful not to show in tables or figures symmetric error bars that would yield results that are out of range (e.g., negative error rates).
        \item If error bars are reported in tables or plots, the authors should explain in the text how they were calculated and reference the corresponding figures or tables in the text.
    \end{itemize}

\item {\bf Experiments compute resources}
    \item[] Question: For each experiment, does the paper provide sufficient information on the computer resources (type of compute workers, memory, time of execution) needed to reproduce the experiments?
    \item[] Answer: \answerYes{} 
    \item[] Justification: The paper reports the computational environment, GPU type and memory, software stack, precision, tensor parallelism, batch-size calibration, and data-reconstruction wall-clock time. Closed models were accessed through the OpenRouter API.
    \item[] Guidelines:
    \begin{itemize}
        \item The answer \answerNA{} means that the paper does not include experiments.
        \item The paper should indicate the type of compute workers CPU or GPU, internal cluster, or cloud provider, including relevant memory and storage.
        \item The paper should provide the amount of compute required for each of the individual experimental runs as well as estimate the total compute. 
        \item The paper should disclose whether the full research project required more compute than the experiments reported in the paper (e.g., preliminary or failed experiments that didn't make it into the paper). 
    \end{itemize}
    
\item {\bf Code of ethics}
    \item[] Question: Does the research conducted in the paper conform, in every respect, with the NeurIPS Code of Ethics \url{https://neurips.cc/public/EthicsGuidelines}?
    \item[] Answer: \answerYes{} 
    \item[] Justification: The research uses public APIs, public prediction-market records, pseudonymous platform identifiers, and no private messages, IP addresses, or real names. The paper also discusses privacy, selection bias, licensing, and responsible release conditions consistent with the NeurIPS Code of Ethics.
    \item[] Guidelines:
    \begin{itemize}
        \item The answer \answerNA{} means that the authors have not reviewed the NeurIPS Code of Ethics.
        \item If the authors answer \answerNo, they should explain the special circumstances that require a deviation from the Code of Ethics.
        \item The authors should make sure to preserve anonymity (e.g., if there is a special consideration due to laws or regulations in their jurisdiction).
    \end{itemize}

\item {\bf Broader impacts}
    \item[] Question: Does the paper discuss both potential positive societal impacts and negative societal impacts of the work performed?
    \item[] Answer: \answerYes{} 
    \item[] Justification: The paper discusses positive impact through reproducible benchmark release and auditing of commitment-grounded language understanding, and discusses negative or risk-relevant aspects, including privacy, pseudonymous user records, selection bias, and limits of interpreting behavioral proxies.
    \item[] Guidelines:
    \begin{itemize}
        \item The answer \answerNA{} means that there is no societal impact of the work performed.
        \item If the authors answer \answerNA{} or \answerNo, they should explain why their work has no societal impact or why the paper does not address societal impact.
        \item Examples of negative societal impacts include potential malicious or unintended uses (e.g., disinformation, generating fake profiles, surveillance), fairness considerations (e.g., deployment of technologies that could make decisions that unfairly impact specific groups), privacy considerations, and security considerations.
        \item The conference expects that many papers will be foundational research and not tied to particular applications, let alone deployments. However, if there is a direct path to any negative applications, the authors should point it out. For example, it is legitimate to point out that an improvement in the quality of generative models could be used to generate Deepfakes for disinformation. On the other hand, it is not needed to point out that a generic algorithm for optimizing neural networks could enable people to train models that generate Deepfakes faster.
        \item The authors should consider possible harms that could arise when the technology is being used as intended and functioning correctly, harms that could arise when the technology is being used as intended but gives incorrect results, and harms following from (intentional or unintentional) misuse of the technology.
        \item If there are negative societal impacts, the authors could also discuss possible mitigation strategies (e.g., gated release of models, providing defenses in addition to attacks, mechanisms for monitoring misuse, mechanisms to monitor how a system learns from feedback over time, improving the efficiency and accessibility of ML).
    \end{itemize}
    
\item {\bf Safeguards}
    \item[] Question: Does the paper describe safeguards that have been put in place for responsible release of data or models that have a high risk for misuse (e.g., pre-trained language models, image generators, or scraped datasets)?
    \item[] Answer: \answerYes{} 
    \item[] Justification: The paper describes safeguards for the released public-data benchmark: only public API data are used, users are represented by platform pseudonyms, and no real names, IP addresses, or private messages are included. The work releases an evaluation benchmark and model outputs rather than a high-risk pretrained model.
    \item[] Guidelines:
    \begin{itemize}
        \item The answer \answerNA{} means that the paper poses no such risks.
        \item Released models that have a high risk for misuse or dual-use should be released with necessary safeguards to allow for controlled use of the model, for example by requiring that users adhere to usage guidelines or restrictions to access the model or implementing safety filters. 
        \item Datasets that have been scraped from the Internet could pose safety risks. The authors should describe how they avoided releasing unsafe images.
        \item We recognize that providing effective safeguards is challenging, and many papers do not require this, but we encourage authors to take this into account and make a best faith effort.
    \end{itemize}

\item {\bf Licenses for existing assets}
    \item[] Question: Are the creators or original owners of assets (e.g., code, data, models), used in the paper, properly credited and are the license and terms of use explicitly mentioned and properly respected?
    \item[] Answer: \answerYes{} 
    \item[] Justification: The paper cites the model and benchmark assets used, and it states the platform terms plus several key licenses.
    \item[] Guidelines:
    \begin{itemize}
        \item The answer \answerNA{} means that the paper does not use existing assets.
        \item The authors should cite the original paper that produced the code package or dataset.
        \item The authors should state which version of the asset is used and, if possible, include a URL.
        \item The name of the license (e.g., CC-BY 4.0) should be included for each asset.
        \item For scraped data from a particular source (e.g., website), the copyright and terms of service of that source should be provided.
        \item If assets are released, the license, copyright information, and terms of use in the package should be provided. For popular datasets, \url{paperswithcode.com/datasets} has curated licenses for some datasets. Their licensing guide can help determine the license of a dataset.
        \item For existing datasets that are re-packaged, both the original license and the license of the derived asset (if it has changed) should be provided.
        \item If this information is not available online, the authors are encouraged to reach out to the asset's creators.
    \end{itemize}

\item {\bf New assets}
    \item[] Question: Are new assets introduced in the paper well documented and is the documentation provided alongside the assets?
    \item[] Answer: \answerYes{} 
    \item[] Justification: The paper introduces and releases StakeBench as a new benchmark asset, documented with data sources, collection funnel, filtering criteria, position reconstruction, task definitions, metrics, prompts, limitations, licensing, and released code/data/model-output packages.
    \item[] Guidelines:
    \begin{itemize}
        \item The answer \answerNA{} means that the paper does not release new assets.
        \item Researchers should communicate the details of the dataset\slash code\slash model as part of their submissions via structured templates. This includes details about training, license, limitations, etc. 
        \item The paper should discuss whether and how consent was obtained from people whose asset is used.
        \item At submission time, remember to anonymize your assets (if applicable). You can either create an anonymized URL or include an anonymized zip file.
    \end{itemize}

\item {\bf Crowdsourcing and research with human subjects}
    \item[] Question: For crowdsourcing experiments and research with human subjects, does the paper include the full text of instructions given to participants and screenshots, if applicable, as well as details about compensation (if any)? 
    \item[] Answer: \answerNA{} 
    \item[] Justification: The paper does not conduct crowdsourcing experiments or direct human-subject studies. It analyzes existing public prediction-market records obtained through public APIs, so there are no participant instructions, screenshots, or compensation details to report.
    \item[] Guidelines:
    \begin{itemize}
        \item The answer \answerNA{} means that the paper does not involve crowdsourcing nor research with human subjects.
        \item Including this information in the supplemental material is fine, but if the main contribution of the paper involves human subjects, then as much detail as possible should be included in the main paper. 
        \item According to the NeurIPS Code of Ethics, workers involved in data collection, curation, or other labor should be paid at least the minimum wage in the country of the data collector. 
    \end{itemize}

\item {\bf Institutional review board (IRB) approvals or equivalent for research with human subjects}
    \item[] Question: Does the paper describe potential risks incurred by study participants, whether such risks were disclosed to the subjects, and whether Institutional Review Board (IRB) approvals (or an equivalent approval/review based on the requirements of your country or institution) were obtained?
    \item[] Answer: \answerNA{} 
    \item[] Justification: The paper does not involve crowdsourcing or direct human-subject experiments; it uses public platform records from public APIs. Accordingly, IRB participant-risk disclosures are not applicable under the manuscript's current study design.
    \item[] Guidelines:
    \begin{itemize}
        \item The answer \answerNA{} means that the paper does not involve crowdsourcing nor research with human subjects.
        \item Depending on the country in which research is conducted, IRB approval (or equivalent) may be required for any human subjects research. If you obtained IRB approval, you should clearly state this in the paper. 
        \item We recognize that the procedures for this may vary significantly between institutions and locations, and we expect authors to adhere to the NeurIPS Code of Ethics and the guidelines for their institution. 
        \item For initial submissions, do not include any information that would break anonymity (if applicable), such as the institution conducting the review.
    \end{itemize}

\item {\bf Declaration of LLM usage}
    \item[] Question: Does the paper describe the usage of LLMs if it is an important, original, or non-standard component of the core methods in this research? Note that if the LLM is used only for writing, editing, or formatting purposes and does \emph{not} impact the core methodology, scientific rigor, or originality of the research, declaration is not required.
    \item[] Answer: \answerYes{} 
    \item[] Justification: LLMs are central evaluated systems in this research, and the paper explicitly declares the 15 evaluated models, access method, decoding settings, prompt templates, label extraction protocol, and model-family comparisons. Any LLM use for writing/editing is not part of the core methodology.
    \item[] Guidelines:
    \begin{itemize}
        \item The answer \answerNA{} means that the core method development in this research does not involve LLMs as any important, original, or non-standard components.
        \item Please refer to our LLM policy in the NeurIPS handbook for what should or should not be described.
    \end{itemize}

\end{enumerate}
\fi

\end{document}